\pdfoutput=1

\documentclass[11pt]{article}
\usepackage[table]{xcolor}

\usepackage[final]{acl}

\usepackage{times}
\usepackage{latexsym}

\usepackage[T1]{fontenc}

\usepackage[utf8]{inputenc}

\usepackage{microtype}

\usepackage{inconsolata}

\usepackage{amsfonts}
\usepackage{amsmath}
\usepackage{graphicx,subfigure}
\usepackage{caption}
\usepackage{subcaption}  
\usepackage{booktabs}
\usepackage{url}
\usepackage{tabularx}
\usepackage{makecell}
\usepackage{comment}
\usepackage{color}
\usepackage{multicol}
\usepackage{multirow}
\usepackage{enumitem}
\usepackage{stfloats}
\usepackage{placeins}
\usepackage{afterpage}
\usepackage{geometry}

\definecolor{en}{HTML}{24A3D6} 
\definecolor{zh}{HTML}{FF5151} 
\definecolor{lang}{HTML}{7AD19C} 

\definecolor{softpastelgreen}{RGB}{198, 233, 211}
\definecolor{softgreen}{RGB}{144,238,144}
\definecolor{darkgreen}{RGB}{0, 100, 0}
\definecolor{mygreen}{RGB}{34,139,34}

\definecolor{softpastelgreen}{RGB}{198, 233, 211}  

\usepackage{tikz}
\usepackage{lscape}          
\usepackage{transparent}
\usepackage{pifont}

\usepackage{kotex} 
\usepackage{CJKutf8}
\usepackage{hyperref}

\newcommand{\algname}{\texttt{MSumBench}} %

\newcolumntype{L}[1]{>{\raggedright\let\newline\\\arraybackslash\hspace{0pt}}m{#1}}
\newcolumntype{X}[1]{>{\centering\let\newline\\\arraybackslash\hspace{0pt}}p{#1}}
\newcolumntype{Y}[1]{>{\raggedleft\let\newline\\\arraybackslash\hspace{0pt}}m{#1}}
\newcolumntype{M}[1]{>{\centering\arraybackslash}m{#1}}

\title{Towards Multi-dimensional Evaluation of LLM Summarization \\across Domains and Languages}

\author{\bf Hyangsuk Min$^{1,}$\thanks{~~Equal Contribution.},\bf ~ Yuho Lee$^{1,*}$,\bf ~ Minjeong Ban$^{1}$,\bf ~ Jiaqi Deng$^{1}$,\\{\bf ~ Nicole Hee-Yeon Kim$^{1}$},{\bf ~ Taewon Yun$^{1}$},{\bf ~ Hang Su$^{2,\dagger}$},{\bf ~ Jason Cai$^{2,}$\thanks{~~This work is conducted independently and is not related to the author(s)' position at Amazon.}~},{\bf ~ Hwanjun Song$^{1}$\thanks{~~Corresponding Author.}}\\
$^{1}$Korea Advanced Institute of Science and Technology\\ $^{2}$AWS AI Labs\\
\ \{hyangsuk.min, yuholee, songhwanjun\}@kaist.ac.kr}

\begin{document}
\maketitle
\begin{abstract}

Evaluation frameworks for text summarization have evolved in terms of both domain coverage and metrics. However, existing benchmarks still lack domain-specific assessment criteria, remain predominantly English-centric, and face challenges with human annotation due to the complexity of reasoning. To address these, we introduce \algname{}, which provides a multi-dimensional, multi-domain evaluation of summarization in English and Chinese. It also incorporates specialized assessment criteria for each domain and leverages a multi-agent debate system to enhance annotation quality. By evaluating eight modern summarization models, we discover distinct performance patterns across domains and languages. We further examine large language models as summary evaluators, analyzing the correlation between their evaluation and summarization capabilities, and uncovering systematic bias in their assessment of self-generated summaries. Our benchmark dataset is publicly available at \href{https://github.com/DISL-Lab/MSumBench}{https://github.com/DISL-Lab/MSumBench}.

\end{abstract}
\begin{figure*}
    \centering
    \includegraphics[width=1\linewidth]{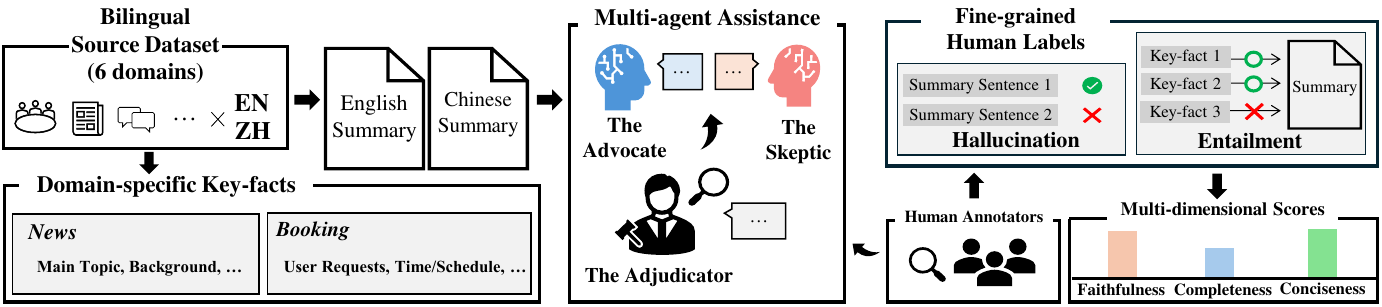}
    \vspace*{-0.6cm}
    \caption{Overview of \algname{}, featuring multi-domain documents in both English and Chinese, with domain-specific key-facts. Model summaries are evaluated via a multi-agent debate framework, aiding annotators' assessments. Each summary then receives percentage scores for faithfulness, completeness, and conciseness.}
    \label{fig:overview}
\vspace*{-0.3cm}
\end{figure*}

\section{Introduction}
\label{sec:introduction}
Recent advancements in large language models (LLMs) have enhanced text summarization performance. However, LLM-generated summaries still face challenges, including hallucinations, omission of critical information, and redundancy \cite{lee2024unisumeval}. These limitations highlight the continued need for advanced automated evaluation methods that can assess summarization quality more efficiently and cost-effectively than human annotation.

While automated evaluations have made progress, there remains a significant gap between automatic evaluations and human judgments, reinforcing the need for benchmarks with high-quality human annotations to provide a more reliable assessment of summarization capabilities.


However, existing benchmarks face three key challenges. First, most benchmarks employ uniform criteria for assessing summary quality, failing to account for domain-specific differences in what constitutes important information \cite{laban2023summedits, lee2024unisumeval}. Second, they remain largely monolingual—primarily focused on English—limiting their ability to provide robust evaluations across languages \cite{bhandari2020realsumm,fabbri2021summeval,pagnoni2021frank,laban2022summac,laban2023summedits,tang2023aggrefact,lee2024unisumeval,tang2024tofueval}.
 Third, collecting high-quality human annotations remains a significant challenge. Assessing summarization quality requires complex reasoning, making the annotation resource-intensive and inconsistent \cite{krishna2023longeval, lee2024unisumeval}. These challenges hinder the timely development of reliable benchmarks that can keep pace with the rapid advancements in LLM capabilities. 

To address these, we create \algname{}, \underline{\textbf{M}}ulti-aspect \underline{\textbf{Sum}}marization \underline{\textbf{Bench}}mark (Figure \ref{fig:overview}), a summarization benchmark for English and Chinese with domain-specific evaluation criteria (see Table \ref{tab:domain_category}).
Building on existing multi-dimensional and fine-grained evaluation frameworks \cite{song2024finesure}, human annotations for summarization quality are collected at both the sentence and key-fact\footnote{A key-fact is a succinct statement capturing an individual essential information unit, containing a maximum of two to three entities \cite{bhandari2020realsumm, song2024finesure}.} levels, focusing on faithfulness, completeness, and conciseness. Furthermore, as depicted in Figure \ref{fig:debate_comparsion}, we propose a multi-agent debate system that facilitates effective AI-human collaboration in handling the complex task of assessing summary faithfulness.
Inspired by prior findings that demonstrate the effectiveness of LLM-based debate \cite{chan2023chateval, du2024multiagentdebate, khan2024debating, koupaee2025faithful}, we extend this approach to the summarization annotation task. Our debate-based system provides annotators with structured arguments from LLM agents--called the Advocate and the Skeptic--with contrasting viewpoints, thereby reducing dependence on any single viewpoint. We further include the Adjudicator’s review to finalize the debate and ensure that all arguments remain fact-based and consistent with the source content. To ease the cognitive load of annotators, we guide them to focus on the most relevant portions of the source text—key entities, relationships, and critical details—rather than requiring them to consider the entire document from scratch. This comprehensive yet focused process not only promotes more accurate annotations but also helps minimize biases from unbalanced or incomplete information.

Our main contributions are: 
(1) We propose an evaluation strategy that identifies different types of critical information specific to each domain. 
(2) We develop a multi-agent, debate-based annotation framework that generates structured arguments with contrasting perspectives, enabling human annotators to focus on key aspects of the task.
(3) We conduct a comprehensive evaluation of state-of-the-art LLMs as summarizers for English and Chinese, using our multi-dimensional annotations. (4) Using the collected annotations as ground truth, we thoroughly examine the effectiveness of LLMs as automated evaluators in both languages. (5) We release the \algname{} benchmark to facilitate advancements in summarization evaluation.

\vspace*{-0.2cm}
\section{Related Works}\label{sec:related-works}
\vspace*{-0.0cm}

\paragraph{Evaluation Benchmarks} 

Conventional summarization benchmarks mainly focus on the news domain \cite{pagnoni2021frank, tang2023aggrefact}, which has led to the development of benchmarks incorporating dialogues \cite{gao2022dialsummeval, krishna2023longeval, tang2024tofueval}, and multi-domains such as SummEdits \cite{laban2023summedits} and UniSumEval \cite{lee2024unisumeval}. However, the monolingual focus on English remains a persistent challenge. While some benchmarks like mFACE \cite{aharoni2023mface} and MFHHD \cite{shen2024mfhhd} provide multilingual evaluations, their focus is limited to the news domain, calling for a single benchmark that considers multiple languages and multiple domains for comprehensive evaluations. Furthermore, while criteria for evaluating summary quality differ across domains, existing multi-domain benchmarks generally employ uniform criteria across different domains \cite{laban2023summedits, lee2024unisumeval}. 

\paragraph{Evaluation Metric}

Many existing benchmarks focus solely on faithfulness (i.e., factual consistency) as an evaluation dimension of summarization quality \citep{bhandari2020realsumm, pagnoni2021frank, laban2022summac, tang2023aggrefact, krishna2023longeval, laban2023summedits}. Some have expanded to multiple dimensions like coherence and relevance \citep{fabbri2021summeval, gao2022dialsummeval, tang2024tofueval}, while \citet{lee2024unisumeval} uses completeness and conciseness for greater evaluation consistency. Evaluation measurement has likewise shifted from coarse, summary-level binary or scale-based ratings \citep{fabbri2021summeval, laban2022summac, gao2022dialsummeval, tang2023aggrefact, laban2023summedits, tang2024tofueval} to fine-grained, sentence-level (or lower) assessments represented as percentage scores \citep{bhandari2020realsumm, pagnoni2021frank, krishna2023longeval, lee2024unisumeval}.

Paralleling these advances, automated metrics have also evolved: from traditional similarity-based measures like ROUGE \cite{lin2004rouge}, BERTScore \cite{yuan2021bartscore}, and BARTScore \cite{zhang2019bertscore}; to natural language inference (NLI) and question answering (QA)-based methods \cite{fabbri2022qafacteval, tang2024minicheck}; and finally to LLM-based approaches \cite{liu2023geval, song2024finesure, wan-etal-2024-acueval}. However, multilingual coverage remains challenging as noted in \citet{forde2024evaluating}. 
Table~\ref{tab:existing_benchmarks} highlights how \algname{} addresses these limitations by comparing it against some of the existing benchmarks.


\begin{table*}[t]
\footnotesize
\renewcommand{\arraystretch}{0.4}
\begin{center}
   \resizebox{\textwidth}{!}{%
\begin{tabular}{ccccccc}
\toprule
\multicolumn{1}{c}{} &
  \multicolumn{1}{c}{\makecell{Domain\\Coverage}} &
  \multicolumn{1}{c}{\makecell{Evaluation\\Dimensions}} &
  \multicolumn{1}{c}{\makecell{Domain-specific\\Evaluation}} &
  \multicolumn{1}{c}{Annotation Unit} &
  \multicolumn{1}{c}{Measurement} &
  \multicolumn{1}{c}{Language} \\
\midrule
\href{https://github.com/google-research/google-research/tree/master/mface}{mFACE} &
  \makecell[c]{\textcolor{red}{\ding{55}} Single} &
  \textcolor{mygreen}{\ding{51}} \textbf{3} &
  \textcolor{red}{\ding{55}} No &
  \makecell[c]{\textcolor{red}{\ding{55}} Summary} &
  \makecell[c]{\textcolor{red}{\ding{55}} Likert} &
  \textcolor{mygreen}{\ding{51}} \textbf{Multi} \\
\href{https://github.com/salesforce/factualNLG}{SummEdits} &
  \textcolor{mygreen}{\ding{51}} \textbf{Multi} &
  \textcolor{red}{\ding{55}} 1 &
  \textcolor{red}{\ding{55}} No &
  \makecell[c]{\textcolor{red}{\ding{55}} Summary} &
  \makecell[c]{\textcolor{red}{\ding{55}} Ternary} &
  \textcolor{red}{\ding{55}} Single \\
\href{https://drive.google.com/drive/folders/1u_7d6dh6FhVwqz5icgYvzp_ltXUM6SlI}{MFHHD} &
  \textcolor{red}{\ding{55}} Single &
  \textcolor{red}{\ding{55}} 1 &
  \textcolor{mygreen}{\ding{51}} \textbf{Yes} &
  \makecell[c]{\textcolor{orange}{\ding{115}} Sentence*} &
  \makecell[c]{\textcolor{red}{\ding{55}} Ternary} &
  \textcolor{mygreen}{\ding{51}} \textbf{Bi} \\
\href{https://github.com/disl-lab/unisumeval-v1.0}{UniSumEval} &
  \textcolor{mygreen}{\ding{51}} \textbf{Multi} &
  \textcolor{mygreen}{\ding{51}} \textbf{3} &
  \textcolor{red}{\ding{55}} No &
  \textcolor{mygreen}{\ding{51}} \textbf{Sentence \& Key-fact} &
  \makecell[c]{\textcolor{mygreen}{\ding{51}} \textbf{Percentage}} &
  \textcolor{red}{\ding{55}} Single \\
\midrule
\textbf{Ours} &
  \textcolor{mygreen}{\ding{51}} \textbf{Multi} &
  \textcolor{mygreen}{\ding{51}} \textbf{3} &
  \textcolor{mygreen}{\ding{51}} \textbf{Yes} &
  \textcolor{mygreen}{\ding{51}} \textbf{Sentence \& Key-fact} &
  \textcolor{mygreen}{\ding{51}} \textbf{Percentage} &
  \textcolor{mygreen}{\ding{51}} \textbf{Bi} \\
\bottomrule
\end{tabular}
}
\vspace*{-0.25cm}
\caption{Benchmark comparison. *The annotated summaries are constrained to a single sentence format.}
\label{tab:existing_benchmarks}
\end{center}
\vspace*{-0.25cm}
\end{table*}

\begin{table*}[ht!]
\footnotesize
\centering
\setlength{\tabcolsep}{2.5pt}
\renewcommand{\arraystretch}{1.0}
\begin{tabular}{cccccc}
\toprule
  News &
  Medical Literature &
  Report &
  Booking &
  Meeting &
  Interview \\
\midrule

  \begin{tabular}[c]{@{}c@{}}Main topic\\ Background\\ Immediate impact\\ Public statements \\ Official statements\\ Counter arguments\\ Future implications\end{tabular} &

  \begin{tabular}[c]{@{}c@{}}Research finding \\ Disease descriptions \\ Medical experiments \\ Medical treatment \\ Medical prevention \end{tabular} &
  \begin{tabular}[c]{@{}c@{}}Governance\\ Evaluations\\ Recommendations\\ Regulation/policy\\ Financial info\end{tabular} &
  \begin{tabular}[c]{@{}c@{}}User requests\\ System suggestions\\ Location/route\\ General information\\ Booking confirmation\\ Price/payment\\ Time/schedule\end{tabular} &
    \begin{tabular}[c]{@{}c@{}}Opinions\\ Reports\\ Decisions\\ Proposals\\ Factual info \end{tabular} &
  \begin{tabular}[c]{@{}c@{}}Background\\ Main arguments\\ Supporting examples\\ Counter arguments\\ Conclusions\end{tabular}\\
  \bottomrule
\end{tabular}
\vspace*{-0.25cm}
\caption{Key-fact categories tailored to each domain, with the first row representing domains and the second row listing domain-specific key-facts. See Appendix \ref{sec:appendix_key-fact} for a detailed description.}
\vspace*{-0.25cm}
\label{tab:domain_category}
\end{table*}

\section{\algname{} Construction Procedure}
\label{sec:pipeline}
We construct \algname{} following the systematic pipeline of four components: dataset collection, domain-specific key-fact generation, summary generation, and summary evaluation. Detailed statistics of \algname{} are provided in Appendix \ref{sec:appendix_dataset}. 

\subsection{Dataset Collection}
\label{sec:input_text_source}
\paragraph{Source Dataset}

Evaluating summarization models on a single domain provides limited insight into the robustness of their performance. Accordingly, \algname{} is constructed based on the datasets from six domains with distinct characteristics: {CNN/DM} (news)\,\cite{nallapati2016abstractive}, {GovReport} (report)\,\cite{huang2021efficient}, {PubMed} (medical literature)\,\cite{cohan2018discourse}, {MultiWOZ} (booking conversation)\,\cite{zang2020multiwoz}, {MediaSum} (interview)\,\cite{zhu2021mediasum}, and {MeetingBank} (meeting) \cite{hu2023meetingbank}. From each domain, we sample 25 documents, yielding a total of 150 source documents to generate summaries.

\paragraph{Documents Translation}
Since \algname{} aims to evaluate both English and Chinese summaries, identical source texts are needed in both languages to ensure contextual consistency and fair comparison. Therefore, we opt to translate the original English source documents into Chinese. The translation follows the three steps. First, we use GPT-4o for initial sentence-level translation with domain-specific prompts (see Appendix~\ref{sec:appendix_translation}) to maintain contextual coherence. Second, we use Qwen-2.5-72B to screen the resulting translations to flag unnatural or inaccurate sentences and to alleviate any bias potentially introduced by using GPT-4o as a single translator. Finally, any flagged sentences are reviewed by bilingual native Chinese examiners for refinement. This multi-step validation ensures the accuracy, naturalness, and contextual integrity of the translations.

\subsection{Domain-Specific Key-Facts Extraction}
Domain-specific criteria are critical for summary evaluation, as each domain emphasizes distinct content. Thus, we define tailored key-fact categories that serve as templates for ideal summary contents in each domain. This approach ensures summary evaluations reflect the most important aspects of original documents while adhering to the established norms and practices of each domain.

\paragraph{Key-Fact Extraction Procedure} The domain-specific key-fact extraction procedure follows three steps. First, \textit{Category Identification}: To determine which information categories matter most in each domain, we analyze human-written reference summaries\footnote{Since MultiWOZ lacks reference summaries, we analyze frequently occurring entity categories labeled by humans.} from existing summarization datasets. Then, we derive approximately 5–7 recurring categories of essential information (see Table~\ref{tab:domain_category}). This ensures that each category captures frequently emphasized details within its respective domain.

Second, \textit{Key-fact Classification}: 
We use GPT-4o to generate candidate key-facts from source documents and classify them using domain-specific categories, filtering out non-aligned key-facts. This minimizes risk of missing any potential key-facts while discarding information that does not meet domain-specific priorities.

Third, \textit{Key-fact Validation}: To enhance reliability, each filtered key-fact is verified by three LLMs—GPT-4o, Claude-3.5-sonnet, and Llama-3.1-70B. Each model checks whether the key-fact: (1) is useful for summarizing; (2) is consistent with the information in the source text; and (3) falls under one of the predefined domain-specific categories. Using a majority vote, any key-fact failing these checks is discarded. This cross-verification mitigates potential biases or oversights that could arise from relying on a single model. The resulting consensus-based key-facts constitute a robust domain-specific reference set, forming the backbone of our subsequent evaluation of summary quality. Appendix~\ref{sec:appendix_key-fact} provides the prompts for extracting and validating the key-facts. 

\subsection{Summary Generation}
To evaluate how summarization capabilities vary across different model scales and architectures--from traditional fine-tuned models to the latest LLMs, we select eight models for benchmarking, categorized into three groups: \emph{non-LLMs}, including fine-tuned BART-Large\,\cite{lewis2020bart} and mT5\,\cite{xue2020mt5}, \emph{open-source LLMs}, including Llama-3.1-70B\,\cite{grattafiori2024llama3herdmodels}, Gemma-2-27b\,\cite{gemmateam2024gemma2improvingopen}, and Qwen-2.5-72B\,\cite{qwen2025qwen25technicalreport}, and \emph{proprietary LLMs}, including GPT-4o\,\cite{openai2024gpt4ocard}, Claude-3.5-Sonnet, and Gemini-1.5-pro\,\cite{geminiteam2024gemini}. 
For English documents, we generate 1,200 summaries (25 source documents × 6 domains × 8 summarizers). For Chinese summaries, we exclude BART due to its lack of multilingual support, resulting in 1,050 generated summaries (25 source documents × 6 domains × 7 summarizers). 
See Appendix~\ref{sec:appendix_summary_generation_detail} for model configuration details.

\subsection{Summary Evaluation}
\label{sec:summary_evaluation}
\paragraph{Human Annotation Tasks} Traditional summary evaluation dimensions (e.g., coherence and relevance) are insufficient for fine-grained evaluation because they lack clear, measurable criteria. Instead, we assess the generated summaries based on three dimensions--faithfulness, completeness, and conciseness--following \citet{song2024finesure}. Our annotation process consists of two main tasks: fact verification and key-fact alignment.



In the fact verification task, annotators evaluate faithfulness at the sentence-level by identifying factual errors based on the existing error taxonomy \cite{lee2024unisumeval} (see Appendix~\ref{sec:appendix_error_type}). In the key-fact alignment task, annotators verify whether each key-fact can be inferred from the summary sentences, similar to an NLI task. These tasks yield 9,951 and 188,800 annotations from each respective task.

We collect our human annotations via Amazon Mechanical Turk (MTurk), with each annotation unit assigned to three independent annotators to ensure reliability. Further details on annotator recruitment and procedure are provided in Appendix~\ref{sec:appendix_annotation}.

\begin{figure}
    \centering
    \includegraphics[width=1\linewidth]{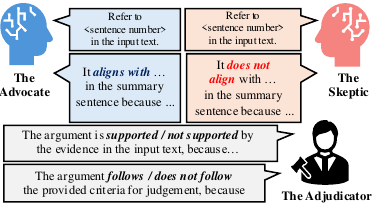}
    \vspace{-0.75cm}
    \caption{Annotation assistance via multi-agent debate.}
    \label{fig:debate_comparsion}
    \vspace{-0.5cm}
\end{figure}

\paragraph{Fact Verification with Multi-Agent Assistance} Although human judgments remain essential for accurate evaluation, the annotation process is both costly and challenging, as summarizers evolve and error types become more nuanced. A straightforward solution might be to have humans inspect labels generated by an LLM-based evaluator, similar to the work of \citet{lee2024unisumeval}. 
However, this can lead to a bias where annotators blindly endorse LLMs' decisions. Thus, we introduce a more systematic approach that fosters effective AI-human collaboration while mitigating bias: a multi-agent debate-assisted annotation framework (Figure~\ref{fig:debate_comparsion}).

Our framework employs three LLM agents for fact verification: the \textit{Advocate}, the \textit{Skeptic}, and the \textit{Adjudicator}. The Advocate and the Skeptic engage in the core debate, while the Adjudicator investigates their arguments. The agents debate the faithfulness of each summary sentence against the full input document (with input sentences numbered).


\begin{table*}[]
\vspace*{-0.0cm}
\centering
\scriptsize
\begin{tabularx}{\textwidth}{c@{\hspace{7pt}}c@{\hspace{7pt}}c@{\hspace{7pt}}c@{\hspace{7pt}}c@{\hspace{7pt}}c@{\hspace{7pt}}c@{\hspace{8pt}}cc@{\hspace{8pt}}c@{\hspace{8pt}}c}
\toprule
\makecell{Annotation Task} & Measuring Dimensions & News & Report & Medical Lit. & Booking & Meeting & Interview & Avg (EN) & Avg (ZH) & Avg (All) \\
\midrule
Fact Verification & Faithfulness & 0.67 & 0.64 & 0.46 & 0.56 & 0.50 & 0.64 & 0.55 & 0.61 & 0.58  \\[3pt]
Key-fact Alignment & Completeness \& Conciseness & 0.70 & 0.84 & 0.77 & 0.83 & 0.80 & 0.82 & 0.77 & 0.82 & 0.79 \\
\bottomrule
\end{tabularx}
\vspace*{-0.25cm}
\caption{Consistency of three human annotators across 6 domains for 2 languages, where the first six columns are the IAA scores for each domains, while the last three are the average IAA for two languages and the overall one.}
\vspace*{-0.25cm}
\label{tab:IAA_main}
\end{table*}

Specifically, the Advocate presents evidence supporting the faithfulness of a given summary sentence, and the Skeptic presents evidence suggesting it may be unfaithful. Both agents first highlight relevant sentence numbers in the input text so annotators can quickly locate the information. Next, they explain which parts of the summary sentence align (or fail to align) with the referenced input sentences (see Table~\ref{tab:prompt_fv_advocate} and \ref{tab:prompt_fv_skeptic}). The Skeptic also specifies the error type. The Adjudicator then reviews both arguments, the input context, and the summary sentence, focusing on two key aspects: (1) whether the provided evidence is consistent with the source text, and (2) whether the arguments follow the faithfulness criteria (see Table~\ref{tab:prompt_fv_adjudicator}). This step prevents superficial objections (e.g., wrongly labeling correct paraphrases as errors). Finally, the Adjudicator produces an investigation report and a tentative label, which are shown to the annotator along with the reference text and both sides’ arguments. We use Llama-3.1-70B for all three agents. 

By providing balanced evidence from both sides and clearly indicating how each argument relates to specific source sentences, this framework minimizes bias and offers a practical aid to annotators. While multi-round debates \cite{ray2023chatgpt, du2024multiagentdebate} may yield richer evaluations, they add cognitive load for annotators, undermining practical benefits as assistance. Conversely, our single-round setup balances thoroughness and usability, ensuring efficient and accurate annotations.

\vspace*{-0.05cm}
\paragraph{Key-Fact Alignment with NLI Assistance}
While the multi-agent debate aids in detecting nuanced faithfulness errors, key-fact alignment is a straightforward comparison between two concise sentences, requiring no complex reasoning mechanisms. 
Therefore, we present annotators with an NLI result generated by Llama-3.1-70B, indicating whether the key-fact is entailed by the summary sentence, along with a brief rationale. This approach streamlines the annotation process by guiding annotators to make quick, confident judgments, thereby reducing their overall cognitive load.

\begin{table}[t!]
\footnotesize
\centering
\setlength{\tabcolsep}{2pt}
\renewcommand{\arraystretch}{0.8}
\begin{tabular}{cccc}
\toprule
Task&Measure & \texttt{UniSumEval} & \algname{}   \\
\midrule
Key-fact List&\begin{tabular}[c]{@{}c@{}}Human Preference\end{tabular} & 
25.00\%    & 
75.00\%       \\
\midrule
Fact Labels&\begin{tabular}[c]{@{}c@{}}Balanced Accuracy\end{tabular}       & 80.07\%    & 92.83\%\\   \bottomrule      
\end{tabular}
\vspace{-0.2cm}
\caption{Comparison of \algname{} over \texttt{UniSumEval}. Human Preference is the A/B test preference ratio, while Balance Accuracy is measured using the two expert examiners' labels as reference.}
\label{tab:unisumeval_comparison}
\vspace{-0.35cm}
\end{table}

\section{Quality Assessment}\label{sec:quality-assessment}


A high-quality benchmark ensures that model evaluations capture genuine performance differences by minimizing annotation noise \cite{lee2024unisumeval, tang2024tofueval}. Thus, we evaluate the reliability and accuracy of collected annotations to ensure the integrity of \algname{}.


\subsection{Annotation Consistency}

Inter-Annotator Agreement (IAA) is a measurement to assess the reliability of human annotations by quantifying the consistency between different annotators.
Table \ref{tab:IAA_main} reports the IAA scores using Krippendorff's $\alpha$ \cite{krippendorff2011computing} across domains for our two annotation tasks, namely {fact verification} and {key-fact alignment}. We achieve very high average IAA scores, Avg (All), of 0.58 and 0.79 for the two tasks, respectively. This indicates that \textbf{MSumBench stands out as the only comprehensive benchmark with domain-specific evaluations and bilingual coverage}, ensuring robust and diverse summarization assessment.

\subsection{Comparison with \texttt{UniSumEval}}
\label{sec:comparsion_uni}

While high IAA scores indicate annotation consistency, they do not guarantee dataset quality due to potential biases and accuracy issues \cite{munappy2022data, braylan2022measuring}. Therefore, we conduct an additional quality check by recruiting two postgraduate NLP specialists as expert examiners, both proficient in English and Chinese (see Appendix~\ref{sec:appendix_annotation} for details on recruitment).

We perform quality checks on two critical components of \algname{}: (1) \emph{Domain-Specific Key-Fact Extraction} to ensure that extracted key-facts align with domain characteristics, and (2) \emph{Multi-Agent Debate} to confirm the accuracy of human annotations assisted by our multi-agent system. To demonstrate the effectiveness of these quality improvements, we compare our dataset with the latest \texttt{UniSumEval} dataset \cite{lee2024unisumeval}.

\begin{table*}[ht]
\scriptsize
\centering
\renewcommand{\arraystretch}{1.4}
\setlength{\tabcolsep}{3.5pt}
\begin{tabular}{ccccccccccc}
\toprule
\multirow{2}{*}{\begin{tabular}[c]{@{}l@{}}Model\\ Type\end{tabular}} & \multirow{2}{*}{Summarizer} & \multicolumn{4}{c}{English} & \multicolumn{4}{c}{Chinese} & \multirow{2}{*}{\begin{tabular}[c]{@{}l@{}}Language\\ Stability\end{tabular}} \\
\cmidrule(lr){3-6} \cmidrule(lr){7-10}
 &  & Faithfulness & Completeness & Conciseness & Domain* & Faithfulness & Completeness & Conciseness & Domain* &  \\
\hline
\multirow{4}{*}{\begin{tabular}[c]{@{}l@{}}Prop- \\rietary \\ LLMs\end{tabular}} 
 & GPT-4o & 
 \cellcolor{en!45}86.10 (5) & 
 \cellcolor{en!55}50.02 (4) & 
 \cellcolor{en!35}77.98 (6) & 
 \cellcolor{en!95}89.28 (1) & 
 \cellcolor{zh!45}78.42 (5) & 
 \cellcolor{zh!75}41.19 (3) & 
 \cellcolor{zh!45}74.98 (5) & 
 \cellcolor{zh!55}86.96 (4) & 
 \cellcolor{lang!55}93.02 (4) \\
 & Claude 3.5 \textsubscript{Sonnet} & 
 \cellcolor{en!95}89.42 (1) & 
 \cellcolor{en!85}52.67 (2) & 
 \cellcolor{en!55}80.09 (4) & 
 \cellcolor{en!85}88.87 (2) &
 \cellcolor{zh!95}82.68 (1) & 
 \cellcolor{zh!85}49.6 (2) & 
 \cellcolor{zh!75}78.07 (3) &
 \cellcolor{zh!75}87.53 (3) & 
 \cellcolor{lang!85}96.3 (2) \\
 & Gemini 1.5 \textsubscript{Pro} &
 \cellcolor{en!55}86.18 (4) & 
 \cellcolor{en!95}54.81 (1) & 
 \cellcolor{en!85}81.88 (2) & 
 \cellcolor{en!55}87.79 (4) & 
 \cellcolor{zh!75}80.26 (3) & 
 \cellcolor{zh!95}53.55 (1) & 
 \cellcolor{zh!85}79.43 (2) & 
 \cellcolor{zh!85}87.55 (2) &
 \cellcolor{lang!95}97.16 (1) \\ \cline{2-11}
 & Average & 
 87.23 (3.33) & 
 52.5 (2.33) & 
 79.98 (4)	& 
 88.65 (2.33)	& 
 80.45 (3) & 	
 48.11 (2)	&  
 77.49 (3.33) & 	
 87.35 (3) & 
95.5 (2.33)\\
\hline
\multirow{4}{*}{\begin{tabular}[c]{@{}l@{}}Open-\\ source\\ LLMs\end{tabular}} 
 & Gemma 2 \textsubscript{27B} & 
 \cellcolor{en!75}87.02 (3) & 
 \cellcolor{en!35}37.06 (6) & 
 \cellcolor{en!15}73.67 (7) & 
 \cellcolor{en!35}84.55 (6) & 
 \cellcolor{zh!55}78.53 (4) & 
 \cellcolor{zh!45}32.18 (5) & 
 \cellcolor{zh!35}69.69 (6) & 
 \cellcolor{zh!35}82.95 (6) &
 \cellcolor{lang!75}93.47 (3) \\
 & Llama 3.1 \textsubscript{70B} & 
 \cellcolor{en!35}83.50 (6) & 
 \cellcolor{en!45}39.19 (5) & 
 \cellcolor{en!75}81.31 (3) & 
 \cellcolor{en!45}87.08 (5) & 
 \cellcolor{zh!35}77.10 (6) & 
 \cellcolor{zh!35}24.78 (6) & 
 \cellcolor{zh!55}75.13 (4) & 
 \cellcolor{zh!95}88.06 (1) & 
 \cellcolor{lang!35}88.41 (6) \\
 & Qwen 2.5 \textsubscript{72B} & 
 \cellcolor{en!85}87.95 (2) & 
 \cellcolor{en!75}50.87 (3) & 
 \cellcolor{en!95}84.87 (1) & 
 \cellcolor{en!75}88.57 (3) & 
 \cellcolor{zh!85}80.29 (2) & 
 \cellcolor{zh!55}40.94 (4) & 
 \cellcolor{zh!95}79.51 (1) & 
 \cellcolor{zh!45}85.81 (5) & 
 \cellcolor{lang!45}92.09 (5) \\ 
 \cline{2-11}
 & Average & 
 86.16 (3.67) & 
 42.37 (4.67)	 & 
 79.95 (3.67)	 & 
 86.73 (4.67)	 & 
 78.64 (4)	 & 
 32.63 (5) & 	
 74.77 (3.67)	 & 
 85.61 (4) & 	
 91.32 (4.67) \\
\hline

\multirow{3}{*}{\begin{tabular}[c]{@{}l@{}}Non-\\ LLMs\end{tabular}} 
 & mT5 & 
 \cellcolor{en!5}12.33 (8) & 
 \cellcolor{en!5}3.79 (8) & 
 \cellcolor{en!5}28.67 (8) & 
 \cellcolor{en!5}55.85 (8) & 
 \cellcolor{zh!15}25.33 (7) & 
 \cellcolor{zh!15}2.86 (7) & 
 \cellcolor{zh!15}27.67 (7) &
 \cellcolor{zh!15}48.76 (7) &
 \cellcolor{lang!15}82.75 (7) \\
 & BART & 
 \cellcolor{en!15}77.57 (7) & 
 \cellcolor{en!15}25.01 (7) & 
 \cellcolor{en!45}78.48 (5) & 
 \cellcolor{en!15}79.2 (7) & - & - & - & - & -\\ \cline{2-11}
 & Average & 
 44.95 (7.5) &	
 14.4 (7.5) & 
 53.58 (6.5)	 & 
 67.53 (7.5) & - & - & - & - & - \\

\bottomrule
\end{tabular}
\vspace*{-0.25cm}
\caption{Summarization performance of eight summarizers for two languages. Rankings are shown in parentheses, with cell color intensity increasing within each column to indicate higher ranks. Domain*: domain stability.}
\label{tab:summ_overall}
\vspace*{-0.25cm}
\end{table*}

\subsubsection{Domain-specific Key-Fact Extraction Quality Check}

Unlike \texttt{UniSumEval}'s domain-agnostic key-fact extraction, \algname{} employs domain-specific categories (Table~\ref{tab:domain_category}) to better reflect each domain's unique characteristics. For comparative evaluation, the two expert examiners perform A/B comparisons between key-facts extracted from 150 {English} source documents using both approaches. 

Table~\ref{tab:unisumeval_comparison} shows that domain-specific key-facts are highly preferred by the examiners. \algname{}'s approach yields a 50\%p higher preference rate than \texttt{UniSumEval} (Cohen's $\kappa$ = 0.48\footnote{Cohen's kappa ($\kappa$) is a statistical measure for inter-rater reliability for two annotators \cite{mchugh2012cohen}.}).
It suggests that \textbf{domain-specific key-facts better capture what humans consider essential in each domain}, leading to more targeted summary evaluation.
{Appendix~\ref{sec:appendix_key-facts_quality_assessment} presents detailed A/B test guidelines and domain-wise results.}



\subsubsection{Multi-Agent Debate Quality Check}

Unlike \texttt{UniSumEval}, which offers a single viewpoint, our debating system mitigates bias by presenting contrasting perspectives. To verify this, we ask the two expert examiners to re-annotate representative samples of summary sentences\footnotemark[4] from both datasets for sentence-level fact verification. The two examiners engage in repeated discussions to reach a consensus on their decisions, ensuring reliable ground-truth labels. 


Table~\ref{tab:unisumeval_comparison} shows that our debating system produces more accurate human labels, achieving a 12.76\%p higher balanced accuracy compared to \texttt{UniSumEval}. {Among incorrect annotations in \texttt{UniSumEval}, 95.31\% align with the single-view assistance label, indicating that the majority of such errors stem from annotators blindly endorsing LLM decisions.}
This reveals that \textbf{a single-view approach inflates IAA scores but harms accuracy, while a multi-view approach can mitigate this by promoting accurate annotation.}

\section{Benchmarking Summarizers}
\label{sec:benchmarking-summarizers}

\paragraph{Dimension and Metric} We evaluate summarization performance across five key dimensions: \emph{faithfulness}, \emph{conciseness}, \emph{completeness}, \emph{domain stability}, and \emph{language stability}. The first three are based on a well-established work \cite{lee2024unisumeval}, and are evaluated as \emph{percentage} scores as suggested by the original paper \cite{song2024finesure}.

Domain and language stability assess the consistency of summarization performance across domains and languages. To assess domain stability, we first compute the coefficient of variation (CV) across domains for each of the first three dimensions-faithfulness, completeness, and conciseness. These three CV values are then averaged to obtain a composite domain stability score. Language stability is assessed similarly, with CVs calculated across languages rather than domains. Detailed calculations are in Appendix ~\ref{sec:appendix_formula}.


\footnotetext[4]{We sample 624 sentences each from \texttt{UniSumEval} (8,133 sentences) and \algname{} (9,951 sentences), achieving 99\% confidence with $\pm$5\% margin of error.}

\subsection{Overview} 

Table \ref{tab:summ_overall} compares the performance of eight summarizers across English and Chinese. Proprietary LLMs demonstrate superior performance compared to open-source and non-LLMs across languages, particularly in completeness. 
They maintain the consistent essential information coverage across languages (52.5 to 48.11), while open-source LLMs experience substantial degradation (42.37 to 32.63). 
This reveals that \textbf{proprietary LLMs capture domain characteristics more effectively in both languages, whereas open-source and non-LLMs do not.} Nevertheless, performance gaps persist across languages regardless of the summarizer type, calling for further efforts to enhance multilingual performance consistency.

\subsection{Detailed Analysis}

We conduct a detailed analysis of the generated summaries, examining completeness and conciseness through key-fact category coverage, and faithfulness through factuality error types.

\begin{figure*}
    \centering
    \includegraphics[width=1\linewidth]{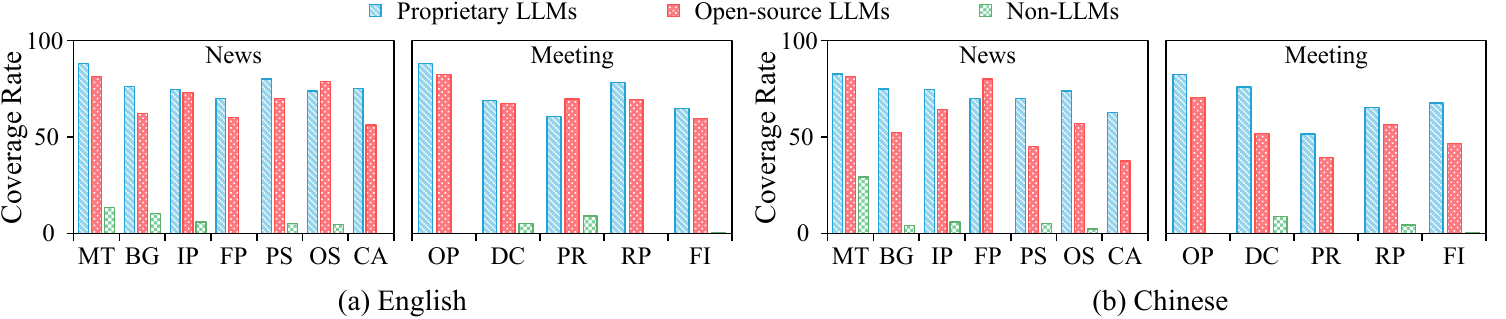}
    \vspace*{-0.75cm}
    \caption{Key-fact category coverage ratio in two domains for English and Chinese. Abbreviations on the x-axis refers: MT(Main topic), BG(Background), IP(Immediate impact), FP(Future implications), PS(Public statements), OS(Official statements), CA(Counterarguments), OP(Opinions), DC(Decisions), PR(Reports), FI(Factual Info).}
    \label{fig:summ_kf}
    \vspace*{-0.4cm}
\end{figure*}

\subsubsection{Comparison on Key-Fact Coverage}

Figure \ref{fig:summ_kf} shows the ratio of key-fact categories covered in the generated summaries. For a detailed analysis, we examine two domains (News and Meeting) in English and Chinese, highlighting coverage differences across three model types.

Generally, in English (Figure \ref{fig:summ_kf}(a)), proprietary and open-source LLMs exhibit relatively consistent key-fact coverage across categories in each domain. However, in Chinese (Figure \ref{fig:summ_kf}(b)), both show greater variability, such as a large gap between MT and CA in News and OP and PR in Meeting. This highlights that \textbf{there is significant category-wise imbalance in key-fact retention in Chinese}, which suggests that similar issues may arise in other non-major languages as well.


From a domain perspective, in general, News exhibits higher coverage rates than Meeting in both languages, though the dominance of key-fact categories in the same domain vary across languages. This reveals that \textbf{all models, regardless of type, still lack satisfactory consistency in capturing key-facts across languages and domains}, which directly impacts the completeness and conciseness of the generated summary.

\subsubsection{Analysis on Factuality Error}

\begin{figure}[t]
    \vspace{0.15cm}
    \centering
    \includegraphics[width=1\linewidth]{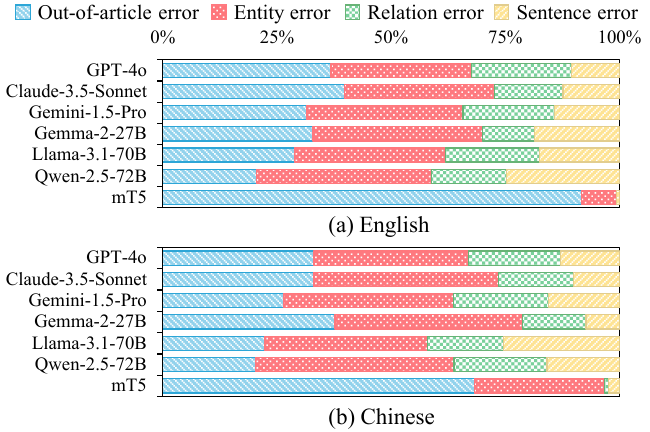}
    \vspace{-0.7cm}
    \caption{Distribution of error types across summarizers and languages. A detailed description of each error type is provided in Appendix \ref{sec:appendix_error_type}.}
    \label{fig:error}
    \vspace{-0.35cm}
\end{figure}

We analyze the distribution of factuality errors appearing in the summaries. Figure \ref{fig:error} shows the distribution across different summarizers for two languages. There is no significant difference in error distribution between proprietary and open-source LLMs, nor across languages. 
The consistency across languages indicates that \textbf{faithfulness in summarization is an inherent property of LLMs rather than a language-dependent factor.} 
Thus, enhancing summarization fidelity is likely to generalize across both models and languages. 

\section{Benchmark: LLMs as Evaluators}
\label{sec:benchmarking-auto-evaluators}

Benchmarking LLMs as summarization evaluators is crucial, as it enables scalable and consistent automated assessment, reducing reliance on costly, time-consuming human evaluation. Their ability to provide objective and reproducible evaluations makes them valuable for benchmarking summarization models across different domains and languages. To assess their effectiveness, we analyze three key aspects:

\smallskip
\noindent$\bullet$ Accuracy of LLMs as evaluators based on their agreement with human judgments.

\smallskip
\noindent$\bullet$ Correlation between LLMs' summarization performance and evaluation performance.

\smallskip
\noindent$\bullet$ Self-evaluation bias, which occurs when an LLM assess summaries it has generated.



\paragraph{Selected Models}
We compare two proprietary LLMs (Claude-3.5-Sonnet, GPT-4o) and two open-source LLMs (Qwen-2.5-72B and Llama-3.1-70B), which exhibits considerably different summarization performance within their respective groups.

\paragraph{Evaluation Metric}
We compare the score obtained by LLM-based evaluation  with human annotated labels in \algname{}. We report the Pearson correlation for the summary-level percentage scores across evaluation dimensions. See Appendix ~\ref{sec:appendix_formula} for the detail of measurements.

\subsection{Correlation with Human Judgments}
\label{sec:autoeval-eval}

Table \ref{tab:autoeval_overall} shows the correlation between LLM and human evaluation across faithfulness, completeness, and conciseness, along with domain and language stability, reflecting correlation consistency across six domains and two languages. 

GPT-4o demonstrates the highest domain and language stability. However, no single model consistently achieves the best performance across all dimensions. Notably, Qwen-2.5-72B stands out for its strong performance in faithfulness evaluation. Therefore, \textbf{there is a significant variability in LLM evaluation accuracy across evaluation dimensions, domains, and languages}, indicating that relying on a single model for automated evaluation lacks reliability.




\begin{table*}[ht!]
\centering
\scriptsize
\renewcommand{\arraystretch}{0.8}
\setlength{\tabcolsep}{3pt}
\begin{tabular}{ccccccccccc}
\toprule
\multirow{2}{*}{\begin{tabular}[c]{@{}c@{}}Model\\ Type\end{tabular}} & \multirow{2}{*}{\begin{tabular}[c]{@{}c@{}}Model\\ Name\end{tabular}} & 
  \multicolumn{4}{c}{English} &
  \multicolumn{4}{c}{Chinese} & \multirow{2}{*}{\begin{tabular}[c]{@{}c@{}}Language\\ Stability \end{tabular}}  \\
  \cmidrule(lr){3-6}\cmidrule(lr){7-10}
 &          & Faithfulness   & Completeness  & Conciseness  & Domain*  & Faithfulness& Completeness  & Conciseness   & Domain*  &       \\
                                   \midrule
\multirow{2}{*}{\begin{tabular}[c]{@{}c@{}}Proprietary\\ LLMs\end{tabular}}  & 
  GPT-4o   & 
  0.61 & 
  \underline{\textbf{0.74}} & 
    0.60 & \underline{\textbf{88.31}} & 0.52 & 0.71 & \underline{\textbf{0.65}} & \underline{\textbf{85.76}} & \underline{\textbf{94.91}} \\
& 

Claude 3.5 \textsubscript{Sonnet}   &
0.60 & 
0.72 & 
\underline{\textbf{0.62}} & 
87.38 & 
0.54 & 
0.65 & 
0.53 & 
80.27& 
90.09 \\
\midrule
\multirow{2}{*}{\begin{tabular}[c]{@{}c@{}}Open-source\\ LLMs\end{tabular}} & Llama 3.1 \textsubscript{70B}  & 0.55  & 0.70  & 0.57  & 85.64  & 0.41  & 0.65  & 0.46  & 75.44  & 93.39 \\
 & Qwen 2.5 \textsubscript{72B}      & \underline{\textbf{0.67}} & 0.65 & 0.55 & 86.96 & \underline{\textbf{0.59}} & \underline{\textbf{0.76}} & 0.55 & 81.11 & 88.40\\
\bottomrule
\end{tabular}
\vspace{-0.25cm}
\caption{Benchmarking LLMs as evaluators in English and Chinese. The higher score, the better accuracy in summary evaluation. Domain*: Domain Stability. Stability scores are computed as in the summarization evaluation.}
\label{tab:autoeval_overall}
\vspace{-0.35cm}
\end{table*}

\begin{figure}[t!]
\centering
\includegraphics[width=1.0\linewidth]{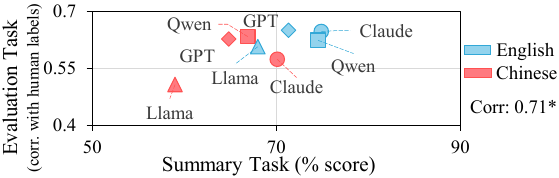}
\vspace{-0.65cm}
\caption{Analysis on cross-task correlation between summarization and evaluation. *: p-value < 0.05.}
\vspace{-0.3cm}
\label{fig:rank_corr}
\end{figure}


\subsection{Cross-Task Correlation}
Figure \ref{fig:rank_corr} shows the correlation between summarization and evaluation performance\footnotemark[5]. 
The x-axis denotes summarization performance, while the y-axis represents its performance as an evaluator.
\footnotetext[5]{Performance here refers to the composite score defined by the average of the three dimensions-faithfulness, completeness, and conciseness.}

We observe \textbf{a strong correlation $\rho$ = 0.71 between LLM performance in summarization and evaluation tasks}, showing that strong summarizers generally excel at evaluation. This suggests that improving an LLM's summarization can also enhance its evaluation, and vice versa. However, as noted in Section \ref{sec:autoeval-eval}, even the best summarizers may still show limitations when evaluating certain dimensions or working with different languages.



\begin{figure}
\centering
\includegraphics[width=1.0\linewidth]{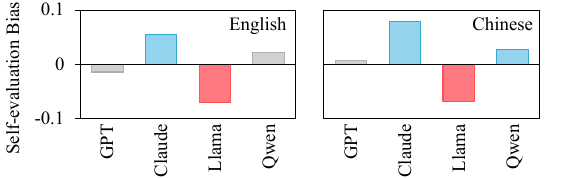}
\vspace*{-0.65cm}
\caption{Self-evaluation bias rates across models and languages. Colored bars denote statistically significant differences determined by t-test (p < 0.05).}
\vspace*{-0.3cm}
\label{fig:self-preference}
\end{figure}

\subsection{Self-Evaluation Bias}


In summary evaluation, LLMs are expected to favor their own summaries \cite{wataoka2024selfpreferencebiasllmasajudge}, but this bias has not been fully analyzed. Thus, we assess to what extent this bias exists. 
We first define the self-evaluation bias rate as the difference between an LLM’s evaluation score for its own summaries and the average score assigned by other evaluators (see Appendix~\ref{sec:appendix_self-preference_bias} for detailed calculation). Thus, the higher (or lower) the rate, the more favorably (or unfavorably) it evaluates its own summaries. We use a t-test to check if the difference is statistically significant.  


Figure~\ref{fig:self-preference} shows the self-evaluation bias rate across LLMs and languages. It reveals that \textbf{self-evaluation bias does not always manifest as self-preference, as it can manifest as either over-rating or under-rating their own work.} Specifically, Llama-3.1-70B rates its own summaries less favorably than others, while Claude-3.5-Sonnet favors its own. This indicates variability in self-assessment across models.

\section{Conclusion}
\label{sec:conclusion}
We create \algname{}, a multi-aspect benchmark for summary evaluation across two languages and six distinct domains. \algname{} extracts key-facts to precisely measure summary-context alignment across six specialized domains. To ensure accurate annotations, we introduce a multi-agent assistant that minimizes human errors and delivers high-quality labels. 
We conduct an in-depth analysis, providing insights into LLMs' behavior as both summarizers and automated evaluators. The open dataset supports evaluation, enhances summarization, and aids preference optimization.
%


\section*{Limitations}\label{sec:lim}

Our work has several limitations as follows:

First, while we incorporate domain-specific key-fact categories, they are derived primarily through statistical patterns. The framework could benefit from direct input by domain experts to capture complex, specialized details more accurately. 

Second, while our evaluation framework accounts for differences in importance across domains, it does not address variations in summary purpose within a single domain. For instance, in medical literature, the focus of a summary could shift from clinical treatments to epidemiological data, depending on summarization perspectives. Future work could explore purpose-specific key-fact categories, aligning summaries more closely with varied summarization objectives. 

Third, our bilingual focus (English and Chinese) is an improvement over monolingual benchmarks. However, it still excludes many underrepresented languages. Evaluating how effectively LLMs handle other lower-resource languages remains an open question.
Lastly, while our multi-agent debate framework provides balanced arguments to assist annotators, exploring other forms of collaboration, such as enhancing debates with dynamic rebuttals, could further improve the effectiveness of AI-human collaboration for summary quality annotation. 
Despite the challenges, we believe that our work serves as a meaningful foundation for future research, particularly in developing more robust automated evaluators and improving the factual consistency of domain-specific summarization systems across different languages.




\section*{Ethics Statement}\label{sec:ethic-statement}

Throughout our study, we prioritized comprehensive communication with all participating annotators, including both crowd-sourced and expert annotators. Crowd-sourced annotators received compensation above the U.S. federal minimum wage rate, while expert examiners were compensated at rates exceeding \$30 per hour, with performance-based incentives. To ensure privacy compliance, all annotator personal information is anonymized in our dataset.

\section*{Scientific Artifacts}\label{sec:sci-artifact}

Our proposed benchmark combines publicly available datasets. For summary generation, we used Huggingface checkpoints and commercial APIs such as OpenAI and AWS Bedrock. Summary model details are in Table ~\ref{tab:summary_model}.



\section*{Acknowledgments}
This work was supported by Institute of Information \& communications Technology Planning \& Evaluation (IITP) grant funded by the Korea goverment (MSIT) (No. RS-2024-00445087, Enhancing AI Model Reliability Through Domain-Specific Automated Value Alignment Assessment).
Additionally, this work was partly supported by the National Research Foundation of Korea (NRF) grant funded by the Korea government (MSIT) (No. RS-2024-00334343).
\bibliography{custom}

\clearpage
\appendix
\newpage
\section{Summary of the Source datasets}
\label{sec:appendix_dataset}
\begin{table*}[htb!]
\scriptsize
\renewcommand{\arraystretch}{1.2}
\setlength{\tabcolsep}{6pt}
\centering
\resizebox{\textwidth}{!}{%
\begin{tabular}{cccccccc}
\toprule
\multirow{2}{*}{Dataset} &
  \multirow{2}{*}{Domain} &
  \multicolumn{2}{c}{English} &
  \multicolumn{2}{c}{Chinse} &
   \\
  \cmidrule(lr){3-4} \cmidrule(lr){5-6}
            &                    & \begin{tabular}[c]{@{}c@{}}Text \\ Word count \\ (Min - Max)\end{tabular}              & \begin{tabular}[c]{@{}c@{}}Summary \\ Word count \\ (Min - Max)\end{tabular}                   & \begin{tabular}[c]{@{}c@{}}Text \\ Word count \\ (Min - Max)\end{tabular}              & \begin{tabular}[c]{@{}c@{}}Summary \\ Word count \\ (Min - Max)\end{tabular}             &       \multirow{2}{*}{\begin{tabular}[c]{@{}c@{}}Key-fact Count\\  (Min - Max)\end{tabular}}          \\
            \midrule
CNN/DM      & News               & 503.6 (234 – 962)    & 86.01 (10 – 205) & 885.4 (447 – 1,657)       &  146.52 (13 – 285)   & 15.72 (6 – 26)  \\
PubMed      & Medical Literature & 2360 (856 – 4496)    & 60.07 (14 – 113) & 3,797.52 (1,272 – 7,862)  & 221.90 (12 – 591)   & 23.84 (10 – 53) \\
GovReport   & Report             & 3401 (1345 – 6837)   & 126.66 (7 – 347) & 6,223.12 (2,466 – 10,359) & 261.71 (10 – 1,704) & 25.08 (16 – 35) \\
MultiWOZ    & Booking            & 243.08 (138 – 382)   & 107.56 (8 – 260) & 465.56 (258 – 788)        & 120.19 (12 – 250)   & 11.52 (6 – 19)  \\
MeetingBank & Meeting            & 547.44 (97 – 1276)   &  71.26 (9 – 143)  & 925.6 (351 – 1,991)       &135.49 (11 – 412)   & 12.64 (5 – 30)  \\
MediaSum    & Interview           & 1108.64 (186 – 4082) & 87.15 (9 – 196)  & 1,951.2 (396 – 7,895)     & 162.47 (15 – 585)   & 16.72 (6 – 27)  \\
\midrule
\multicolumn{2}{c}{\algname{}}         & 1,360 (138 – 6,837)  & 2374.73 (258-10,359)      & 89.89 (7 - 347)  & 174.71 (10 – 1,704) & 17.59 (5 – 53) \\
\bottomrule
\end{tabular}
}
\caption{Overview of six datasets utilized in 
 \algname{}: includes statistics on source documents and their summaries, showing mean word counts and key-fact quantities, along with their respective ranges (min to max values).}
\label{tab:dataset_specification}
\end{table*}

Table~\ref{tab:dataset_specification} presents a comprehensive analysis of six diverse datasets encompassing various domains, including news, medical literature, reports, booking, meetings, and interviews in both English and Chinese languages. These datasets are strategically selected to evaluate the model's capability in handling domain-specific contextual understanding. The evaluation benchmark consists of 150 source documents, evenly distributed with 25 documents per domain in each language.

\section{Dataset Construction Pipeline Detail}
\label{sec:appendix_pipeline}

\begin{table}[hbt!]
\footnotesize
\centering
\setlength{\tabcolsep}{6pt}
\begin{tabular}{cccc}
\toprule
Domain &
  \begin{tabular}[c]{@{}c@{}}Retained \\ Key-Facts\end{tabular} &
  \begin{tabular}[c]{@{}c@{}}Newly Added \\ Key-Facts\end{tabular} &
  \begin{tabular}[c]{@{}c@{}}Removed \\ Key-Facts\end{tabular} \\
  \midrule
News              & 63.6 \% & 36.4\% & 14.0 \% \\
Medical Lit. & 58.1 \% & 41.9 \% & 24.9 \% \\
Report            & 60.8 \% & 39.2 \% & 23.1 \% \\
Booking           & 79.2 \% & 20.8 \% & 4.8 \%  \\
Meeting           & 72.2 \% & 27.8 \% & 17.4 \% \\
Interview         & 70.6 \% & 29.4 \% & 16.7 \% \\
Overall           & 67.4 \% & 32.6 \% & 16.8 \%\\
\bottomrule
\end{tabular}
\caption{Key-fact comparison between domain-specific method (A) and generic methods (B): Retained Key-facts represent the percentage of key-facts that remain consistent between A and B. Newly Added Key-facts indicate the percentage of key-facts introduced in A, but absent in B. Removed Key-facts show the percentage of key-facts present in B but missing in A.}
\label{tab:key-fact_comparsion}
\end{table}

\begin{table}[hbt!]
\footnotesize
\setlength{\tabcolsep}{4pt}
\begin{tabular}{cccc}
\toprule
Domain &  \texttt{UniSumEval} & \algname{} &  Cohen's kappa\\
\midrule
News & 32.0\%    & 68.0 \% & 0.46          \\
Medical Lit. & 10.0 \%    & 90.0 \% & 0.34          \\
Report             & 14.0 \%    & 86.0 \% & 0.50          \\
Booking            & 22.0 \%    & 78.0 \% & 0.43          \\
Meeting            & 38.0 \%    & 62.0 \% & 0.43          \\
Interview          & 26.0 \%    & 74.0 \% & 0.48          \\
\midrule
Overall            & 21.7 \%    & 78.3 \% & 0.47      \\
\bottomrule
\end{tabular}
\vspace{-0.2cm}
\caption{The percentage scores of A/B testing for domain-specific key-facts human preferences between \texttt{UniSumEval} and \algname{}.}
\label{tab:key-fact_preference}
\end{table}

\begin{table*}[hbt!]
\setlength{\tabcolsep}{12pt}
\footnotesize
\centering
\begin{tabular}{ccc}
\toprule
Model Name        & \begin{tabular}[c]{@{}c@{}}Hugging Face Checkpoints \\ \& Official API Version \end{tabular}    &    Hardware \& Precision \\
\midrule
GPT-4o            & gpt-4o-2024-08-06    &  API (OpenAI, default settings)             \\
Claude-3.5-Sonnet & anthropic.claude-3-5-sonnet-20241022-v2:0 & API (AWS Bedrock, default settings)          \\
Gemini-1.5-Pro    & gemini-1.5-pro-002       & API (Google API, default settings)           \\
Gemma-2-27B       & google/gemma-2-27b-it     & NVIDIA L40S 48GB (×2) \& BF16        \\
Llama-3.1-70B     & meta-llama/Llama-3.1-70B-Instruct  & NVIDIA L40S 48GB (×4) \& BF16\\
Qwen2.5-72B       & Qwen/Qwen2.5-72B-Instruct         & NVIDIA L40S 48GB (×4) \& BF16\\
mT5               & sebuetnlp/mT5\_multilingual\_XLSum & NVIDIA L40S 48GB (×1) \& Full precision \\
BART & \begin{tabular}[c]{@{}c@{}}facebook/bart-large-cnn\\ linydub/bart-large-samsum\end{tabular} & \begin{tabular}[c]{@{}c@{}}NVIDIA L40S 48GB (×1) \& Full precision\end{tabular}\\
\bottomrule
\end{tabular}%
\caption{The checkpoint of the model used to generate summaries.}
\vspace*{-0.3cm}
\label{tab:summary_model}
\end{table*}

\begin{table}[hbt!]
\footnotesize
\setlength{\tabcolsep}{3pt}
\begin{tabular}{cccc}
\toprule
Domain &  Direct Copy(\%) & \begin{tabular}[c]{@{}c@{}}Near-Exact\\ Match(\%) \end{tabular} &  \begin{tabular}[c]{@{}c@{}}Extractiveness\\ Score(\%) \end{tabular}\\
\midrule
News & 4.68\%   & 16.31\%   & 0.63         \\
Medical Lit. & 0.19\%   & 18.10\%   & 0.64          \\
Report             & 5.68\%   & 23.52\%   & 0.68          \\
Booking            & 4.96\%   & 3.34\%   & 0.52          \\
Meeting            & 4.67\%   & 13.88\%   & 0.58          \\
Interview          & 3.08\%   & 9.33\%   & 0.53          \\
\midrule
Overall            & 3.88\%   & 14.08\%   & 0.60      \\
\bottomrule
\end{tabular}
\vspace{-0.2cm}
\caption{The measurements of extractiveness of key-fact across domains in three ways: direct copy, near-exact match and extractiveness score.}
\vspace{-0.6cm}
\label{tab:key-fact_extractivness}
\end{table}

\subsection{Translation Detail}
\label{sec:appendix_translation}
\begin{table*}[ht]
        \footnotesize
    \centering
\resizebox{\textwidth}{!}{%
\begin{tabular}{p{0.9\linewidth}} 
    \toprule
You are an expert English-Chinese translator specialized in news articles. Your task is to translate the following English text to Chinese (Simplified Chinese), sentence by sentence, with careful attention to quality and accuracy. \\ [3pt] \\ [3pt]

Warning: Use only "standard Simplified Chinese characters" and English technical terms when necessary \\ [3pt] \\ [3pt]

Translation Rules:
\begin{enumerate}
    \item  Reference Consistency
    \begin{itemize}
        \item Keep organization names in original form
        \item Translate ALL PERSON NAMES to Chinese following appropriate conventions:
        \begin{itemize}
            \item Western names: Use standard Chinese transliteration
            \begin{itemize}
                \item Example: Michael → \begin{CJK}{UTF8}{gbsn}迈克尔 (Màikè'ěr)\end{CJK}, John → \begin{CJK}{UTF8}{gbsn}约翰 (Yuēhàn)\end{CJK}
            \end{itemize}
            \item Chinese names: Maintain Chinese characters
            \begin{itemize}
                \item Keep family name and given name format (e.g., \begin{CJK}{UTF8}{gbsn}王小明\end{CJK})
            \end{itemize}
            \item For established figures, use their commonly known Chinese name
            \begin{itemize}
                \item Example: Shakespeare → \begin{CJK}{UTF8}{gbsn}莎士比亚 (Shāshìbǐyà)\end{CJK}
            \end{itemize}
        \end{itemize}
        \item Use standard format for dates, times, and numbers
    \end{itemize}

 \item Technical Terms
 \begin{itemize}
     \item Use established Chinese technical terms
     \item First mention: Chinese term (English term); Following mentions: Chinese term only
     \item Maintain consistency in specialized terms throughout
 \end{itemize}

 \item Cultural Adaptation
 \begin{itemize}
     \item Translate English idioms and proverbs to Chinese cultural equivalents (\begin{CJK}{UTF8}{gbsn}成语\end{CJK} when appropriate)
     \item Convert Western business expressions to match Chinese business etiquette: 
     \begin{itemize}
         \item Use appropriate level of formality (\begin{CJK}{UTF8}{gbsn}敬语\end{CJK})
         \item Follow Chinese business conversation conventions
     \end{itemize}
     \item Maintain neutrality and objectivity in expression
     \item Add brief explanations for culturally specific items
 \end{itemize}

 \item  Chinese Writing Style Consistency
 \begin{itemize}
     \item Use formal written Chinese (\begin{CJK}{UTF8}{gbsn}书面语\end{CJK}) consistently
     \item Avoid mixing formal and colloquial expressions
     \item Follow standard news writing conventions:
     \begin{itemize}
         \item Use proper \begin{CJK}{UTF8}{gbsn}判断词\end{CJK} and \begin{CJK}{UTF8}{gbsn}状态词\end{CJK}
         \item Use standard news article punctuation
     \end{itemize}
     \item Word choice guidelines: 
     \begin{itemize}
         \item Prefer \begin{CJK}{UTF8}{gbsn}因为\end{CJK} over \begin{CJK}{UTF8}{gbsn}由于\end{CJK} for causation
         \item Use \begin{CJK}{UTF8}{gbsn}表示\end{CJK} instead of \begin{CJK}{UTF8}{gbsn}说\end{CJK} for formal statements
        \item Choose \begin{CJK}{UTF8}{gbsn}认为\end{CJK} over \begin{CJK}{UTF8}{gbsn}觉得\end{CJK} for opinions
     \end{itemize}
 \end{itemize}
\end{enumerate}

Provide your answer ONLY in this simple JSON format: \\ [3pt]
\{"translation": ["First Chinese translation", "Second Chinese translation", "Third Chinese translation", ..., "Last Chinese translation"]\}\\ [3pt] \\ [3pt]

English text: \{input\_text\} \\ [3pt]
    \bottomrule 
    \end{tabular}
    }
    \caption{Chinese translation prompt for news domain.}
    \label{tab:prompt_translation_nondial}
\end{table*}
\begin{table*}[ht]
    \footnotesize
    \centering
\resizebox{\textwidth}{!}{%
\begin{tabular}{p{0.9\linewidth}} 
    \toprule
You are an expert English-Chinese translator specialized in medical and scientific literature. Your task is to translate the following English text to Chinese (Simplified Chinese), sentence by sentence, with careful attention to quality and accuracy. \\ [3pt] \\ [3pt]

Warning: Use only "standard Simplified Chinese characters" and English technical terms when necessary \\ [3pt] \\ [3pt]

Translation Rules:
\begin{enumerate}
    \item  Medical Terminology
    \begin{itemize}
        \item Use standardized Chinese medical terms (\begin{CJK}{UTF8}{gbsn}规范医学用语\end{CJK})
        \item Keep precision in medical concepts:
        \begin{itemize}
            \item Diseases: Standard Chinese names (\begin{CJK}{UTF8}{gbsn}英文名\end{CJK})
            \item Medications: Generic names in Chinese (\begin{CJK}{UTF8}{gbsn}英文通用名\end{CJK})
            \item Medical procedures: Standard translations
        \end{itemize}
        \item Handle technical terms:
        \begin{itemize}
            \item First mention: Chinese term (English term); Following mentions: Chinese term only
        \end{itemize}
        \item  Maintain absolute consistency in terminology throughout
        \item Abbreviations: Keep standardized format
        \begin{itemize}
            \item First mention: Full Chinese term (Full English term [Abbreviation]) (\begin{CJK}{UTF8}{gbsn}聚合酶链式反应\end{CJK}: Polymerase Chain Reaction [PCR])
        \end{itemize}
        \item Special terms:
        \begin{itemize}
            \item Gene/protein names: Follow standard conventions
            \item Chemical formulae: Maintain accuracy
            \item Anatomical terms: Use standard translations
        \end{itemize}
    \end{itemize}
    \item Academic Writing Style
    \begin{itemize}
        \item Use formal academic Chinese (\begin{CJK}{UTF8}{gbsn}学术用语\end{CJK}) 
        \item Follow scientific writing conventions: use precise and objective language and maintain the scientific tone
        \item Sentence structure: Clear and concise, logical flow, one key point per sentence
        \item Word choice guidelines:
        \begin{itemize}
            \item Use \begin{CJK}{UTF8}{gbsn}发现\end{CJK} for findings
            \item  Use \begin{CJK}{UTF8}{gbsn}显示\end{CJK} for results presentation
            \item  Use \begin{CJK}{UTF8}{gbsn}表明\end{CJK} for conclusions
            \item  Use \begin{CJK}{UTF8}{gbsn}证实\end{CJK} for verification
            \item  Use \begin{CJK}{UTF8}{gbsn}提示\end{CJK} for implications
            \item  Use \begin{CJK}{UTF8}{gbsn}比较\end{CJK} for comparison
            \item  Use \begin{CJK}{UTF8}{gbsn}分析\end{CJK} for analysis
        \end{itemize}
    \end{itemize}
\end{enumerate}

    Provide your answer ONLY in this simple JSON format: \\ [3pt]
    \{"translation": ["First Chinese translation", "Second Chinese translation", "Third Chinese translation", ..., "Last Chinese translation"]\}\\ [3pt] \\ [3pt]
    
    English text: \{input\_text\} \\ [3pt]
    \bottomrule 
    \end{tabular}
    }
    \caption{Chinese translation prompt for medical literature domain.}
    \label{tab:prompt_translation_nondial_pubmed}
\end{table*}
\begin{table*}[ht]
    \footnotesize
    \centering
    \resizebox{\textwidth}{!}{%
    \begin{tabular}{p{0.9\linewidth}} 
    \toprule
You are an expert English-Chinese translator specialized in reports and official documents. Your task is to translate the following English text to Chinese (Simplified Chinese), sentence by sentence, with careful attention to quality and accuracy. \\ [3pt] \\ [3pt]

Warning: Use only "standard Simplified Chinese characters" and English technical terms when necessary \\ [3pt] \\ [3pt]

Translation Rules:
\begin{enumerate}
    \item  Reference Consistency
    \begin{itemize}
        \item Keep organization names in original form
        \item Translate ALL PERSON NAMES to Chinese following appropriate conventions:
        \begin{itemize}
            \item Western names: Use standard Chinese transliteration (Michael → \begin{CJK}{UTF8}{gbsn}迈克尔 (Màikè'ěr)\end{CJK})
            \item Chinese names: Maintain Chinese characters
        \end{itemize}
        \item For established figures, use their commonly known Chinese name (Shakespeare → \begin{CJK}{UTF8}{gbsn}莎士比亚 (Shāshìbǐyà)\end{CJK})
        \item Use standard Chinese format for:
        \begin{itemize}
            \item Dates: \begin{CJK}{UTF8}{gbsn}YYYY年MM月DD日\end{CJK}
            \item Numbers: Use Chinese numerals for formal documents (\begin{CJK}{UTF8}{gbsn}一、二、三\end{CJK})...)
            \item Percentages: Use \begin{CJK}{UTF8}{gbsn}百分之\end{CJK} format
            \item Monetary values: Follow Chinese currency notation
        \end{itemize}
    \end{itemize}

    \item Technical Terms
    \begin{itemize}
        \item Use established Chinese technical terms
        \item First mention: Chinese term (English term); Following mentions: Chinese term only
        \item Maintain consistency in specialized terms throughout
    \end{itemize}
    \item Cultural Adaptation
    \begin{itemize}
        \item Translate English idioms and proverbs to Chinese cultural equivalents (\begin{CJK}{UTF8}{gbsn}成语\end{CJK} when appropriate)
        \item Convert Western business expressions to match Chinese business etiquette:
        \begin{itemize}
            \item Use appropriate level of formality (\begin{CJK}{UTF8}{gbsn}敬语\end{CJK})
            \item Follow Chinese business conversation conventions
        \end{itemize}
        \item Maintain neutrality and objectivity in expression
        \item Add brief explanations for culturally specific items
    \end{itemize}

    \item Chinese Writing Style Consistency
    \begin{itemize}
        \item Use the highest level formal written Chinese (\begin{CJK}{UTF8}{gbsn}政府公文体\end{CJK})
        \item Follow official document writing conventions:
        \begin{itemize}
            \item Use standard official vocabulary (\begin{CJK}{UTF8}{gbsn}规范用语\end{CJK})
            \item Apply proper ceremonial words (\begin{CJK}{UTF8}{gbsn}礼仪用语\end{CJK})
        \end{itemize}
        \item Use formal written Chinese (\begin{CJK}{UTF8}{gbsn}书面语\end{CJK}) consistently
        \item Avoid mixing formal and colloquial expressions
        \item Sentence structure:
        \begin{itemize}
            \item Use parallel structure for lists (\begin{CJK}{UTF8}{gbsn}排比句\end{CJK})
            \item Maintain appropriate formality level
            \item Use standard government document punctuation
        \end{itemize}
    \end{itemize}
\end{enumerate}

    Provide your answer ONLY in this simple JSON format: \\ [3pt]
    \{"translation": ["First Chinese translation", "Second Chinese translation", "Third Chinese translation", ..., "Last Chinese translation"]\}\\ [3pt] \\ [3pt]
    
    English text: \{input\_text\} \\ [3pt]
    \bottomrule 
    \end{tabular}%
    }
    \caption{Chinese translation prompt for report domain.}
    \label{tab:prompt_translation_nondial_govreport}
\end{table*}

\begin{table*}[ht]
    \footnotesize
    \centering
\resizebox{\textwidth}{!}{%
\begin{tabular}{p{0.9\linewidth}} 
    \toprule
You are an expert English-Chinese translator specialized in dialogue scripts. Your task is to translate the following English script to Chinese (Simplified Chinese), sentence by sentence, with careful attention to quality and accuracy.\\ [3pt] \\ [3pt]

Warning: Use only "standard Simplified Chinese characters" and English technical terms when necessary. \\ [3pt] \\ [3pt]

Translation Rules:
\begin{enumerate}
    \item Reference Consistency
    \begin{itemize}
        \item Keep organization names in original form
        \item Translate ALL personal names to Chinese following appropriate conventions:
        \begin{itemize}
            \item Western names: Use Chinese transliteration (James → \begin{CJK}{UTF8}{gbsn}詹姆斯\end{CJK})
            \item Chinese names: Keep Chinese characters
            \item For established figures, use their commonly known Chinese name (Shakespeare →  \begin{CJK}{UTF8}{gbsn}莎士比亚 (Shāshìbǐyà)\end{CJK})
        \end{itemize}
        \item The same Chinese translation must be used when referring to that person within any dialogue.
    \end{itemize}

\item  Technical Terms
\begin{itemize}
    \item Use established Chinese technical terms
    \item First mention: Chinese term (English term); Following mentions: Chinese term only
\end{itemize}

\item Speaking Style \& Format
\begin{itemize}
    \item Keep "Speaker: dialogue" format 
    \begin{itemize}
        \item Place actions in parentheses
        \item Maintain conversation flow
    \end{itemize}

     \item Use appropriate formal Chinese based on context
     \item Keep each speaker's tone consistent
     \item se proper conversational particles (\begin{CJK}{UTF8}{gbsn}吧，呢，啊\end{CJK})
     \item Adapt greetings and courtesies to Chinese norms(\begin{CJK}{UTF8}{gbsn}您好，请问，麻烦您\end{CJK})
\end{itemize}
\end{enumerate}

Provide your answer ONLY in this simple JSON format: \\ [3pt]

\{"translation": ["First Chinese translation", "Second Chinese translation", "Third Chinese translation", ..., "Last Chinese translation"]\} \\ [3pt] 

English text: \{input\_text\} \\ [3pt]
        \bottomrule 
    \end{tabular}
    }
    \caption{Chinese translation prompt for booking/meeting/interview domain}
    \label{tab:prompt_translation_dial}
\end{table*}
\begin{table*}[ht]
    \footnotesize
    \centering
    \resizebox{\textwidth}{!}{%
    \begin{tabular}{p{0.9\linewidth}} 
    \toprule
You are an expert English-Chinese translator specialized in handling precise information across various domains. Your task is to translate the following English key facts into Chinese (Simplified Chinese), sentence by sentence, with careful attention to quality, accuracy and consistency. Read the given document carefully, fully understand it, and keep that in mind when you translate key fact sentences. Use all terms already written in Chinese from the document to ensure consistency across the translation. Follow the instructions to translate English key fact sentences. \\ [3pt] \\ [3pt]

Warning: Use only "standard Simplified Chinese characters" and English technical terms when necessary.\\ [3pt]

Translation and Verification Rules: \\ [3pt]
\begin{enumerate}
    \item Reference Consistency
    \begin{itemize}
        \item Keep organization names in original form
        \item  Translate ALL PERSON NAMES to Chinese following appropriate conventions:
        \begin{itemize}
            \item  Western names: Use standard Chinese transliteration
            \begin{itemize}
                \item Example: "Michael" → \begin{CJK}{UTF8}{gbsn}"迈克尔" (Màikè'ěr), "John" → "约翰" (Yuēhàn)\end{CJK}
            \end{itemize} 
            \item Chinese names: Maintain Chinese characters
             \begin{itemize}
                \item Keep family name and given name format (e.g., \begin{CJK}{UTF8}{gbsn}王小明\end{CJK})
             \end{itemize}
        \end{itemize}
    \item  For established figures, use their commonly known Chinese name
    \begin{itemize}
        \item Example: Shakespeare → \begin{CJK}{UTF8}{gbsn}莎士比亚 (Shāshìbǐyà)\end{CJK}
    \end{itemize}
    \item  Use standard format for dates, times, and numbers
    \end{itemize}

\item Technical Terms
\begin{itemize}
    \item  Use established Chinese technical terms
    \item  Follow terms already defined in the document for consistency.
    \item  First mention: Chinese term (English term); Following mentions: Chinese term only
    \item  Maintain consistency in specialized terms throughout
\end{itemize}

\item Focus on Information
\begin{itemize}
    \item   Prioritize the accurate transfer of factual information in each key fact.
    \item   Avoid any stylistic adjustments or embellishments. Translate the text plainly and faithfully.
\end{itemize}
\item Back-Translation for Verification:
\begin{itemize}
    \item For each translated Chinese sentence:
    \begin{itemize}
        \item Perform a back-translation into English.
        \item Compare the back-translation with the original English key fact.
    \end{itemize}
    \item If there is any difference in meaning, revise the Chinese translation and repeat Steps 1–3 until the back-translation aligns with the original English sentence.
\end{itemize}
\end{enumerate}

Provide your answer in JSON format: \\ [3pt]
{{"translation": [("1", "Chinese translation"), ("2", "Chinese translation"), ..., ("Key fact number", "Chinese translation")]}} \\ [3pt] \\ [3pt]

Document:  \\ [3pt]
\{input\_text\}  \\ [3pt]

\{N\} Key fact sentences: \\ [3pt]
\{key\_facts\} \\
    \bottomrule 
    \end{tabular}
    }
    \caption{Chinese translation prompt for key-fact.}
    \label{tab:prompt_translation_key-fact}
\end{table*}
\begin{table*}[h]
    \footnotesize
    \centering
\resizebox{\textwidth}{!}{%
\begin{tabular}{p{0.9\linewidth}} 
    \toprule
 You are an expert English-Chinese translator with extensive experience in translation quality assessment. Your task is to check the quality of the English-Chinese translation, sentence by sentence, with careful attention to quality and accuracy. \\[3pt]

Quality check instructions:\\[3pt]
\begin{enumerate}
    \item Accuracy
    \begin{itemize}
        \item Compare the English source text and Chinese translation to ensure meaning is preserved
        \item Check for any omissions or additions
        \item Verify numerical values, dates, and proper names are correctly translated
        \item Flag any mistranslations or semantic errors
    \end{itemize}

    \item Consistency
    \begin{itemize}
        \item Reference Consistency: Check if proper nouns, organization names, and product names are translated consistently
        \item Technical Term Consistency: Verify industry-specific terminology is translated consistently and correctly
        \item Style Consistency: Ensure consistent tone and level of formality throughout
    \end{itemize}

    \item Fluency
    \begin{itemize}
        \item Check if the translation reads naturally in Chinese
        \item Verify proper Chinese grammar and syntax
        \item Ensure appropriate sentence structure and flow
        \item Check for any awkward expressions or unnatural phrasing
    \end{itemize}
    \item Readability
    \begin{itemize}
        \item Assess if the text is easy to understand for the target audience
        \item Check sentence length and complexity
        \item Verify proper paragraph breaks and text organization
        \item Ensure clear logical flow
    \end{itemize}

\item Cultural Appropriateness
\begin{itemize}
    \item Check for cultural sensitivity
    \item Verify idioms and expressions are appropriately localized
    \item Ensure measurements, dates, and currencies are properly converted
    \item Flag any potential cultural misunderstandings
\end{itemize}

\item Professionalism
\begin{itemize}
    \item Verify appropriate formal/business language usage
    \item Check for proper honorific forms
    \item Ensure professional terminology is correctly used
    \item Maintain appropriate level of formality
\end{itemize}

\item Fitness for Purpose
\begin{itemize}
    \item Verify the translation meets its intended purpose
    \item Check if appropriate for target audience
    \item Ensure industry-specific requirements are met
    \item Verify technical accuracy for specialized content
\end{itemize}

\end{enumerate}

Provide the answer using the following JSON format: \\ [3pt]
\{"translation": [(1, "True", "IF True Blank", "IF True Blank", "IF True Blank"), (2, "False", "IF False English Sentence", "IF False Chinese Sentence", "IF False Reason"), ..., (N, "True", "", "", "")]
\}  \\ [3pt]

English text: \{input\_text\}  \\ [3pt]
Chinese text: \{translation\_text\}  \\ [3pt]
    \bottomrule 
    \end{tabular}
    }
    \caption{Translation qaulity check prompt.}
    \label{tab:prompt_translation_quality_check}
\end{table*}
We present implementation details for the translation of source documents. For English-Chinese translation of source documents and key-facts, we incorporate domain-specific writing styles into our prompts, as shown in Tables~\ref{tab:prompt_translation_nondial} -~\ref{tab:prompt_translation_dial}. 
The key-fact translation prompt used for the key-fact alignment task with auto-evaluators is presented in Table~\ref{tab:prompt_translation_key-fact}. 
To ensure translation quality, we developed a screening prompt detailed in Table~\ref{tab:prompt_translation_quality_check}, which evaluates translations sentence-by-sentence across multiple dimensions: accuracy, consistency, fluency, comprehensibility, cultural adaptation, formality, and adequacy. 

\subsection{Key-Fact Generation}
\label{sec:appendix_key-fact}

\begin{table*}[]
\footnotesize
\renewcommand{\arraystretch}{1.5}
\setlength{\tabcolsep}{2pt}
\centering
\begin{tabular}{lll}
\toprule
\multicolumn{1}{c}{Domain} &
  \multicolumn{1}{c}{Category} &
  \multicolumn{1}{c}{Description} \\
\midrule
\multirow{7}{*}{News} &
  Main Topic &
  General information about the primary event, issue, or occurrence being discussed \\
 &
  Background &
  Situational details supporting the main topic \\
 &
  Immediate Impact &
  Short-term effects or consequences resulting from the main topic \\
 &
  Future Implications &
  Long-term outcomes or projected developments related to the main topic \\
 &
  Public statements &
  Non-expert perspectives, opinions, or reactions from the general public \\
 &
  Official statements &
  Expert or authoritative opinions, assessments, or analysis on the main topic \\
 &
  Counterarguments &
  Critiques or opposition to the main topic or its impacts \\
\midrule
\multirow{5}{*}{\begin{tabular}[c]{@{}l@{}}Medical \\ Literature\end{tabular}} &
  Research Finding &
  Key discoveries or outcomes from medical research studies \\
 &
   Medical experiments &
  {\begin{tabular}[c]{@{}l@{}} Detailed procedures, methodologies, and designs of experiments \\or clinical trials testing treatments or interventions\end{tabular}} \\
 &
  Disease descriptions &
  Detailed explanations of diseases, including symptoms, causes, and characteristics \\
 &
  Medical Treatment &
  Recommended therapies, treatments, and interventions aimed at managing or curing diseases \\
 &
  Medical Prevention &
  Strategies and actions aimed at preventing diseases and promoting public health \\
\midrule
\multirow{5}{*}{Report} &
  Recommendations &
  Suggested actions or improvements based on report findings \\
 &
  Governance &
  Oversight and administration of programs and resources \\
 &
  Regulation and Policy &
  Legal frameworks, standards, and policies guiding operations \\
 &
  Evaluations &
  Assessment of data and review of program performance \\
 &
  Financial information &
  Details about costs, budget allocations, and financial impacts \\
\midrule
\multirow{7}{*}{Booking} &
  General Information &
  Reference numbers, contact info, headcounts, or else excluding time, location, and price \\
 &
  Price and Payment &
  Cost, pricing, fees and payment methods \\
 &
  Time and Schedule &
  Time slots, schedules, and booking times \\
 &
  Location and Route &
  Address, location and routes \\
 &
  Booking Confirmation &
  Confirmation and status of bookings for the services \\
 &
  User Requests &
  User's requests for the booking service or details \\
 &
  System Suggestions &
  Suggestions provided by the system. \\
\midrule
\multirow{5}{*}{Meeting} &
  Opinions &
  Personal viewpoints, perspectives, or feelings regarding a topic or proposal \\
 &
   Decisions &
  {\begin{tabular}[c]{@{}l@{}} specific, actionable plans or choices made to address a Problem \\ or improve a situation after careful consideration\end{tabular}}  \\
 &
   Proposals &
  Final conclusion or choice made after discussion, determining the course of action \\
 &
   Reports &
 {\begin{tabular}[c]{@{}l@{}}  Structured updates or presentations summarizing status, data, \\or findings related to projects or objectives\end{tabular}}  \\
 &
   Factual Information &
  {\begin{tabular}[c]{@{}l@{}}  Objective data, statistics, or verified information \\ that serves as a basis for decisions or discussions\end{tabular}} \\
\midrule
\multirow{5}{*}{Interview} &
  Background &
  Context or historical info to help understanding \\
 &
   Main Arguments &
  Core claims or opinions from each speaker \\
 &
   Supporting Examples &
  Examples, data, or statistics to support the main arguments \\
 &
  Counterarguments &
  Opposing views or criticisms of the main arguments and responses \\
 &
  Conclusions &
  Key points and future directions from the interview \\
\bottomrule
\end{tabular}
\caption{Description of key-fact category for each domain.}
\label{tab:key-fact_description}
\end{table*}
\begin{table*}[]
\footnotesize
\renewcommand{\arraystretch}{1.5}
\setlength{\tabcolsep}{15pt}
\centering
\begin{tabular}{lll}
\toprule
\multicolumn{1}{c}{Domain} & \multicolumn{1}{c}{Category} & \multicolumn{1}{c}{Example}                                           \\
\midrule
\multirow{7}{*}{News}      & Main Topic                   & Prince William is scheduled at the Cenotaph.                          \\
                           & Background                & Guy Thorpe-Beeston has 18 years’ experience.                          \\
                           & Immediate Impact          & William will travel by car, not helicopter.                           \\
                           & Future Implications       & The delivery room at St Mary’s Lindo Wing will have a new team.       \\
                           & Public statements         & Prosecutor Brice Robin confirmed no videos were used.                 \\
                           & Official statements       & Obstetrician Guy Thorpe-Beeston will lead the birth.                  \\
                           & Counterarguments          & Prince William faces a conflict between duties and his child's birth. \\
\midrule
\multirow{5}{*}{\begin{tabular}[c]{@{}l@{}}Medical \\ Literature\end{tabular}} & Research Finding & Intervention strategies have helped reduce malaria globally. \\
                           & Medical experiments       & New mapping techniques track malaria transmission.                    \\
                           & Disease descriptions      & Dengue is caused by four viruses (DENV 1-4).                          \\
                           & Medical Treatment         & Dengue treatment involves haematological monitoring.                  \\
                           & Medical Prevention        & Insecticide-treated bednets have been key to malaria control.         \\
\midrule
\multirow{5}{*}{Report}    & Recommendations           & HHS agreed to gather more disenrollment data.                         \\
                           & Governance                & CMS reviews aspects of contracts between states and D-SNPs.           \\
                           & Regulation and Policy     & Dual-eligible beneficiaries qualify for Medicare and Medicaid.        \\
                           & Evaluations               & CMS lacks data on disenrolled beneficiaries.                          \\
                           &  Financial information     & Medicaid covers premiums for dual-eligible beneficiaries.             \\
\midrule
\multirow{7}{*}{Booking}   & General Information       & The train ticket number is 12345.                                     \\
                           & Price and Payment         & The total cost for the train ticket is \$45.                          \\
                           & Time and Schedule         & The restaurant is available from 6 PM to 9 PM.                        \\
                           & Location and Route        & The Grand Hotel is at 123 Kings Road, Brighton, BN1 2GS.              \\
                           & Booking Confirmation      & A table for 8 is booked.                                              \\
                           & User Requests             & The user wants to book a hotel room for two people.                   \\
                           & System Suggestions        & The system recommends taking a train to Cambridge.                    \\
\midrule
\multirow{5}{*}{Meeting}   & Opinions                  & Dave Shukla argues consolidation increases expenses.                  \\
                           & Decisions                 & The council decided to consolidate gas under one commission.          \\
                           & Proposals                 & Dave Shukla proposes keeping utilities independent.                   \\
                           & Reports                   & The goal is to create one commission for cost savings.                \\
                           & Factual Information       & Reid concludes the party’s divisiveness cost them control.            \\
                           \midrule
\multirow{5}{*}{Interview} & Background                & The discussion covers Obama’s State of the Union.                     \\
                           & Main Arguments            & Cary says establishment Republicans struggle against Trump.           \\
                           & Supporting Examples       & Cary cites Republican grassroots efforts in early states.             \\
                           & Counterarguments          & Cary defends the Republican Party, claiming not all are hateful.      \\
                           & Conclusions               & Reid concludes the party’s divisiveness cost them control.           \\
\bottomrule
\end{tabular}
\caption{Example of key-fact category for six domains.}
\label{tab:key-fact_example}
\end{table*}

\begin{table*}[h]
    \scriptsize
    \centering
 \resizebox{\textwidth}{!}{%
 \begin{tabular}{p{0.9\linewidth}} 
    \toprule
    Your task is to identify domain-specific key facts within the document in <document></document>, which are essential pieces of information for a high-quality summary. You are provided key-fact category and description in <category></category> tags, and its example in <example></example>. Each key fact must be presented as a standalone, atomic-level sentence. The following is a set of detailed instructions in <instructions></instructions> tags for identifying key facts. \\ [3pt] \\ [3pt]
<instructions>
\begin{enumerate}
    \item Identify Key-facts: Extract all key-facts from the text. Each key-fact should:
    \begin{itemize}
        \item Be a complete sentence with a subject, verb, and object/complement.
        \item Contain only one action, event, or idea. Avoid compound sentences.
        \item Include no more than two or three entities per key fact. If there are more than three entities, divide them into separate sentences.
        \item Present temporal information (e.g., when, how long) as standalone sentences.
        \item Present causal relationships (e.g., reasons, consequences) as standalone sentences.
        \item Avoid using linking words like 'and,' 'but,' 'then,' or 'because.' Each idea must be presented as an independent fact.
        \item Do not combine related details, even implicitly. Each sentence must describe exactly one action, idea, or relationship.
    \end{itemize}
    \item Here are the examples of key-fact structure granularity: 
    \begin{itemize}
        \item Text Example: The resolution authorizes the operation of one winter shelter from December 1st, 2019, to March 31st, 2020.
        \item  Key-facts (Revised): 
        \begin{enumerate}
         \item The resolution authorizes the operation of one winter shelter.
         \item The winter shelter will operate from December 1st, 2019, to March 31st, 2020. 
        \end{enumerate}
         \item Text Example : The property for the winter shelter was purchased using homeless emergency aid program funding.
         \item Key-facts (Revised):  
         \begin{enumerate}
             \item The property for the winter shelter was purchased. 
             \item Homeless emergency aid program funding was used for the purchase. 
         \end{enumerate}
    \end{itemize}
    \item Categorize Key-facts: 
    \begin{itemize}
    \item[--] Define your own categories for the key-facts based on content.
    \item[--]  Assign each key-fact to a category.
    \end{itemize}
   
\item Compare Categories: 
\begin{itemize}
    \item[--] Compare your defined categories with the provided key-fact categories. 
    \item[--] Adjust any key-fact to better align with the provided categories.
\end{itemize}
\item Validate Key-facts
\begin{itemize}
    \item Ensure each key-fact meets the following criteria:
    \begin{itemize}
        \item Correctly categorized.
        \item Atomicity: Conveys only one action, event, or idea.
        \item  Clarity: Is concise and clear, avoiding ambiguity.
        \item  Brevity: Contains no unnecessary details.
        \item  Non-overlapping: Does not duplicate information in other facts.
    \end{itemize}
    \item Finalize key facts that satisfy these conditions.
\end{itemize}
\end{enumerate}
</instructions> \\ [3pt] \\ [3pt]
<category> \\ [3pt]
Here are the provided Key Fact Categories and Descriptions: \\ [3pt]
{categories} \\ [3pt]
</category> \\ [3pt] \\ [3pt]

<example> \\ [3pt]
Here are the examples of key facts to illustrate the level of granularity for each category: \\ [3pt]
\{category\_examples\} \\ [3pt]
</example> \\ [3pt] \\ [3pt]

Provide your answer in JSON format. The answer should be a dictionary with tuples: \\ [3pt]
'key\_facts' containing the key facts as a list of tuples (key\-fact, reason, category):  \\ [3pt]
\{"key\_facts": [ ("first key fact", "reason", "category"), ("second key fact", "reason", "category") ]\} \\ [3pt]
<document> \\ [3pt]
\{input\_text\} \\ [3pt]
</document> \\
    \bottomrule
    \end{tabular}
    }
    \caption{Domain-specific key-fact generation prompt.}
    \label{tab:prompt_key-fact_generation}
\end{table*}
\begin{table*}[h]
        \footnotesize
    \centering
\resizebox{\textwidth}{!}{%
\begin{tabular}{p{0.9\linewidth}} 
    \toprule
    You will receive a Document and a set of Key-facts sentences that contain essential pieces of information from the Source Document. Your task is to identify if each Key-Fact Sentence is useful for making a summary of the Source Document.\\ [3pt]  \\ [3pt]
    
Reasons a Key-Fact Sentence may NOT be useful: \\ [3pt]
Reason 1) Trivial Information: The sentence contains correct but insignificant details that do not contribute meaningfully to the main points of the Source Document.
\begin{itemize}
    \vspace*{-0.3cm}
    \item Example: The Source Document is a business meeting transcript, but the Key-Fact Sentence is "Speaker A said 'Good afternoon.'"
    \vspace*{-0.3cm}
    \item Explanation: While accurate, this detail doesn't enhance understanding of the meeting's objectives or outcomes.
\end{itemize}

Reason 2) Incorrect Information: The sentence includes factual errors or discrepancies compared to the Source Document.
\begin{itemize}
    \vspace*{-0.3cm}
   \item Example: The Source Document mentions 12 children living in a small town, but the Key-Fact Sentence states 11 children in a big city.
    \vspace*{-0.3cm}
    \item  Explanation: The numbers and locations don't match, making the sentence inaccurate.
\end{itemize}
Reason 3) Irrelevant Information: The sentence is unrelated to the content of the Source Document.
\begin{itemize}
    \vspace*{-0.3cm}
    \item Example: The Source Document is about artificial intelligence, but the Key-Fact Sentence discusses a dog barking loudly.
    \vspace*{-0.3cm}
    \item Explanation: The sentence doesn't pertain to the topic of the Source Document.
\end{itemize}
Reason 4) Category Alignment: The sentence is incorrectly categorized.
\begin{itemize}
    \vspace*{-0.3cm}
    \item Example: A sentence about future plans is categorized as "Factual Information" when it should be "Proposals"
    \vspace*{-0.3cm}
    \item Explanation: The content of the sentence doesn't match the characteristics of its assigned category.
\end{itemize}
Reason 5) Domain Relevance: The information in the sentence, while accurate and properly categorized, is not essential for summarizing this type of document.
\begin{itemize}
    \vspace*{-0.3cm}
    \item Example: In a meeting minutes document discussing a major company merger, a sentence about routine office maintenance schedule categorizing as "Factual Information"
    \vspace*{-0.3cm}
    \item Explanation: Although this information is accurate and properly categorized as "Factual Information", it's not essential for understanding the key points of a merger discussion meeting.
\end{itemize}

Here are the Instructions for Key-facts validation:
\begin{enumerate}
    \vspace*{-0.3cm}
    \item Read a Document and a set of Key-facts sentences carefully.
    \vspace*{-0.3cm}
    \item Evaluate each Key-facts sentence based on the five reasons above.
    \vspace*{-0.3cm}
    \item According to evaluation, if the Key-facts sentence is useful for making a summary of the Source Document, response "Yes", otherwise response "No"
    \vspace*{-0.3cm}
    \item Provide a single sentence explaining why the Key-facts sentence is useful for making a summary.
\end{enumerate}

Provide your answer in JSON format. The answer should be a list of dictionaries whose keys are "sentence", "response", "reason". \\ [3pt]
you should provide a response and a reason for all Key-facts sentence: [\{"sentence": "first key-fact sentence", "response": "your response", "reason": "your reason"\}, \{"sentence": "second key-fact sentence", "response": "your response", "reason": "your reason"\}, ... , \{"sentence": "N-th key-fact sentence", "response": "your response", "reason": "your reason"\}] \\ [3pt]

Document:  \\ [3pt]
\{input\_text\}  \\ [3pt]  \\ [3pt]
\{N\} Key-facts sentence with Category:  \\ [3pt]
\{key\_fact\}\\
    \bottomrule 
    \end{tabular}
    }
    \caption{Key-fact validation prompt.}
    \label{tab:prompt_key-fact_validation}
\end{table*}
\paragraph{Key-Fact Extraction and Validation}
Table~\ref{tab:prompt_key-fact_generation} presents the domain-specific key-facts generation prompt, including well-defined categories and detailed domain-specific descriptions, along with representative examples illustrating granularity. The categories for each domain are detailed in Table~\ref{tab:key-fact_description}, with corresponding examples in Table~\ref{tab:key-fact_example}. To ensure the quality of extracted key-facts, we perform key-fact validation using the prompt in Table~\ref{tab:prompt_key-fact_validation}.

\vspace{-0.3cm}
\paragraph{Key-Fact Extractiveness} We provide additional analysis on the extractiveness of key-facts, in relation to their source documents. Since key-facts are designed to function as atomic information units, analyzing their extractiveness helps characterize their structural properties. We measure extractiveness using three metrics. First, direct copy measures the percentage of key-facts that exactly match a sentence in the source document. Second, near-exact match captures the proportion of key-facts with over 90\% similarity to the source, as measured by the Edit Distance algorithm. Such cases typically involve surface-level edits, including whitespace changes or removal of redundant adjectives. Third, we compute the extractiveness score using the metric proposed by~\citet{song2024finesure}, defined as the average n-gram overlap $(n=1/3/5)$ between each key-fact and the source document.

Table~\ref{tab:key-fact_extractivness} shows the extractiveness patterns across domains. Direct verbatim copying is relatively rare, with an average rate of 3.88\%. Near-exact matches average 14.08\%, indicating that most key-facts undergo some degree of reformulation from the original text. The extractiveness score averages 0.60, showing moderate lexical overlap with source documents, consistent with the interpretation by \citet{song2023enhancing}. These results indicate that key-facts are typically reformulated to some extent, rather than directly copied, reflecting their role as processed, atomic information units suitable for systematic evaluation.

\subsection{Summary Generation Detail}
\label{sec:appendix_summary_generation_detail}
\begin{table*}[h]
\footnotesize
    \centering
  \begin{tabular}{p{0.95\linewidth}} 
    \toprule
    (a) English summary generation prompt \\
    \midrule
    Text: \{input\_text\} \\[5pt]  \\
    Instruction: Summarize the Text. \\[5pt]  \\
    Provide your answer in JSON format: The answer should be a dictionary with the key "summary" containing a generated summary as a string:\\ 
    \{"summary": "your summary"\} \\  
    \midrule
    (b) Chinese summary generation prompt \\
    \midrule
    \begin{CJK}{UTF8}{gbsn}文本: \{input\_text\}\end{CJK}  \\[5pt]  \\
    \begin{CJK}{UTF8}{gbsn}指令: 请用中文总结这段文字。\end{CJK}\\[5pt]  \\
    \begin{CJK}{UTF8}{gbsn}以 JSON 格式提供答案。答案应是一个包含 "summary" 键和值的字典，值为生成的摘要字符串：\end{CJK}\\
    \begin{CJK}{UTF8}{gbsn}\{"summary": "总结内容"\}\end{CJK} \\
    \bottomrule
    \end{tabular}
    \caption{Summary generation prompt.}
    \label{tab:prompt_summary}
\end{table*}

Table~\ref{tab:summary_model} details the model versions employed for summary generation. For non-LLMs, we utilized pre-trained models from Hugging Face, while for open-source LLMs, we implemented instruction-tuned checkpoints. Access to proprietary LLMs was facilitated through their respective official APIs. The prompts for both English and Chinese summary generation are detailed in Table~\ref{tab:prompt_summary}, with the temperature parameter set to 1.0 to promote diversity in generated summaries.

\subsection{Key-Fact Alignment Detail}
\label{sec:appendix_key-fact_alignment}
\begin{table*}[h]
    \footnotesize
    \centering
\begin{tabular}{p{0.95\linewidth}} 
    \toprule
    You will be provided with two sets of information:
    \begin{itemize}
        \item[--]List of sentences A: A list of sentences that need to be evaluated.
        \item[--]sentence B: A sentence that will be checked against sentence A. 
    \end{itemize}
Your task is to evaluate whether the complete information in each sentence from List A is fully contained in sentence B. This requires performing n evaluations (where n is the number of sentences in List A). For example, if there are 15 sentences in List A, you have to do 15 evaluations. \\ [3pt]  \\ [3pt]
 Instruction:\\ [3pt]
 \begin{itemize}
     \item [--]A sentence from List A is considered "contained" in the sentence B if and only if:
     \begin{enumerate}
         \item The complete information conveyed by the sentence in List A is entirely present in the sentence B.
         \item The essential meaning of the sentence in List A must align with or be fully and explicitly captured by sentence B.
         \item  The information conveyed by the sentence in List A must be explicitly implied or fully understood by sentence B.
     \end{enumerate}
     \item Exact wording is not required, but the complete and explicit meaning must match.
     \item Provide a short reason, and a label: contained, not contained.
 \end{itemize}

Please provide your answer in JSON format:\\ [3pt]
[\{"sentence A": "1", "label": "contained", "reason": "Short explanation of why you chose this label"\}, \{"sentence A": "sentence A number", "label": "contained or not contained", "reason": "Short explanation of why you chose this label"\}, ..., \{"sentence A": "N", "label": "not contained", "reason": "Short explanation of why you chose this label"\}]

Note:\\[3pt]
\begin{itemize}
    \item [--]Ensure that all n evaluations are performed without omission. Each combination of a sentence from List A and the sentence B must be explicitly evaluated.
    \item [--]Any skipped or missing evaluations will result in incomplete analysis, so please confirm that no sentence in List A is overlooked.
    \item [--]Clearly explain your reasoning even if the label is "not contained."
    \item [--]Ensure your evaluation strictly adheres to the requirement that the entire meaning of the sentence in List A should be fully present in sentence B to be labeled as "contained."
    \item [--]Criteria recap: contained, not contained
\end{itemize}
    
List of sentences A: \\ [3pt]
\{key-fact\} \\ [3pt]\\ [3pt]

Sentence B:\\ [3pt]
\{summary\}\\ [3pt]\\ [3pt]
    \bottomrule 
    \end{tabular}
    \caption{Key-fact alignment prompt.}
    \label{tab:prompt_key-fact_alignment}
\end{table*}
Table~\ref{tab:prompt_key-fact_alignment} presents our streamlined NLI prompt for key-fact alignment assessment. The prompt instructs the model to perform a direct entailment check between summary sentences and key-facts, providing brief, focused explanations. This design enables quick and confident assessment of information alignment while minimizing unnecessary complexity.

\subsection{Multi-Agent Debate System Detail}
\label{sec:appendix_multi-agent}
The prompts used for our multi-agent fact verification system are presented in Tables~\ref{tab:prompt_fv_advocate} -~\ref{tab:prompt_fv_adjudicator}. Each agent's prompt is tailored to their specific role: the Advocate focuses on finding supporting evidence, the Skeptic emphasizes identifying potential contradictions or flaws, and the Adjudicator concentrates on analyzing both arguments against the source document for final verification.
\begin{table*}[h]
    \scriptsize
    \centering
    \begin{tabular}{p{0.9\linewidth}} 
    \toprule
    You are the ADVOCATE, an agent defending the factual consistency of the summary. Assume the summary sentences are always true and faithful. Cite specific sentences from the reference document as evidence to support your claim for each summary sentence. You are given the reference document provided in <document></document> tags and the summary sentences in <summary></summary> tags. Critically assess and present your reasoning. \\ [3pt] \\ [3pt]
    <errors> 
    \vspace{-0.2cm}
    \begin{itemize}
        \item out-of-article error: If the summary sentence introduces facts, subjective opinions, or biases that cannot be verified or confirmed by the reference document, the summary is factually inconsistent with the reference document. 
        \vspace{-0.2cm}
        \item entity error: If the summary sentence includes incorrect or misrepresented entities, such as names, numbers, or main subjects, the summary is factually inconsistent with the reference document.
        \vspace{-0.2cm}
        \item relation error: If the summary sentence contains incorrect semantic relationships, such as the use of wrong verbs, prepositions, or adjectives, which distort the relationships between entities, the summary sentence is factually inconsistent with the reference document. 
        \vspace{-0.2cm}
        \item sentence error: If the summary sentence completely contradicts the information in the reference document, requiring significant revision or removal to align with the reference document, the summary sentence is factually inconsistent with the reference document.
    \end{itemize}
    \vspace{-0.2cm}
    </errors>  \\ [3pt] \\ [3pt]
    <instructions>  \\ [3pt]
    Here is the instructions for writing your arguments: \\ [3pt]
    Follow these steps carefully to ensure a structured and thorough evaluation under your assigned role.  \\ [3pt]
    
    1. Read the reference document and summary sentence under your role:  \\ [3pt]
    \vspace{-0.5cm}
    \begin{itemize}
        \item[--] Carefully read the reference document and try to fully understand it. 
        \vspace{-0.2cm}
        \item[--] Compare each summary sentence to the reference document to identify evidence supporting its factual consistency. 
        \item[--] Refer to the error types in <errors></errors> tags to better defend your claim (faithfulness), focusing on areas of alignment and acceptable variations.
        \vspace{-0.2cm}
    \end{itemize}
    2. As a ADVOCATE, focus on finding alignment:  \\ [3pt]
    \vspace{-0.3cm}
    \begin{itemize}
        \item[--] Explicitly identify numbered sentences in the reference document that support or partially align with the summary sentence. 
        \vspace{-0.2cm}
        \item[--] Even if a perfect match cannot be found, select the closest sentence(s) that contain key elements (entities, relationships, events, quantities, or cause-effect relationships).
        \vspace{-0.2cm}
        \item[--] Always select at least one numbered sentence from the reference document, even if it partially aligns with the summary.  
        \vspace{-0.2cm}
        \item[--] As an ADVOCATE, focus on defending the faithfulness of the summary while addressing any potential inconsistencies.
    \end{itemize}
    \vspace{-0.2cm}    
    3. Provide a detailed explanation of your arguments:  \\ [5pt]
    \vspace{-0.5cm}
    \begin{itemize}
        \item[--]  For each summary sentence:
        \vspace{-0.2cm}
        \begin{itemize}
            \item[$\bullet$] Cite one or more numbered sentences from the reference document, even if only partial alignment exists. 
            \item[$\bullet$] Use the format "reference\_sentence\_number": [number1, number2] to explicitly indicate the reference of support.
            \item[$\bullet$] Include a "reason" that explicitly explains how the cited sentences align with the summary, focusing on entities,  relationships, and key details.
        \end{itemize}
        \vspace{-0.2cm}
        \item[--] Your explanation must:
        \vspace{-0.2cm}
        \begin{itemize}
            \item[$\bullet$] Be concise.
            \item[$\bullet$] Reference specific elements, such as: 
            \begin{itemize}
                \item Key details: Validate quantities, events, and cause-effect relationships. 
                \item Rephrasing accuracy: Argue that the summary retains the original meaning despite rephrasing.
                \item Word choice: Ensure terms like "only," "significant," or "most" match the intensity or scale of the reference document.
            \end{itemize}
        \end{itemize}
    \end{itemize}
    </instructions>
    
    Provide your answers in JSON format as shown below:  \\ [1pt]
    [\{ "summary\_sentence\_num": "0", "label": "1 ( if faithfulness 1, else 0 )", "error\_type": "no error ( if faithfulness no error )",  \\ [1pt]
        "reference\_sentence": ["sentence from reference document", "another sentence from reference document"], \\ [1pt]
        "reference\_sentence\_number": [0, 1], \\ [1pt]
        "reason": "The summary mentions the entity 'X', but the reference document refers to 'Y' [0]. Additionally, the relationship between 'A' and 'B' is misrepresented, as the reference indicates 'A causes B', not 'A results from B' [1]."\}, \\ [1pt]
      \{"summary\_sentence\_num": "N", "label": "1 ( if faithfulness 1, else 0 )",         "error\_type": "no error ( if faithfulness no error )",  \\ [1pt]
        "reference\_sentence": ["closest available sentence"], \\ [1pt]
        "reference\_sentence\_number": [10], \\ [1pt]
        "reason": "The summary states 'Event Z happened', but no reference to 'Event Z' is found in the reference [10]." \}] \\ [1pt] \\ [1pt]
    
    Reference document, divided into numbered sentences:  \\ [3pt]
    <document>  \\ [3pt]
    {input\_text} \\ [3pt]
    </document> \\ [3pt] \\ [3pt]
    
    Summary with {N} sentences: \\ [3pt]
    <summary> \\ [3pt]
    {summaries} \\ [3pt]
    </summary> \\ [3pt]
    \bottomrule
    \end{tabular}
    \caption{Fact verification ADVOCATE prompt.}
    \label{tab:prompt_fv_advocate}
\end{table*}
\begin{table*}[h]
    \scriptsize
    \centering
\resizebox{\textwidth}{!}{%
    \begin{tabular}{p{0.9\linewidth}} 
    \toprule
    You are the SKEPTIC, an agent identifying and arguing for factual inconsistencies in the summary, considering error types. Assume the summary sentences are always unfaithful. Cite specific sentences from the reference document as evidence to support your claim for each summary sentence. You are given the reference document provided in <document></document> tags and the summary sentences in <summary></summary> tags. Now, follow the instructions in <instructions></instructions> tags. Critically analyze and present your reasoning. \\ [3pt] \\ [3pt]
    <errors> 
    \vspace{-0.2cm}
    \begin{itemize}
        \item out-of-article error: If the summary sentence introduces facts, subjective opinions, or biases that cannot be verified or confirmed by the reference document, the summary is factually inconsistent with the reference document. 
        \vspace{-0.2cm}
        \item entity error: If the summary sentence includes incorrect or misrepresented entities, such as names, numbers, or main subjects, the summary is factually inconsistent with the reference document.
        \vspace{-0.2cm}
        \item relation error: If the summary sentence contains incorrect semantic relationships, such as the use of wrong verbs, prepositions, or adjectives, which distort the relationships between entities, the summary sentence is factually inconsistent with the reference document. 
        \vspace{-0.2cm}
        \item sentence error: If the summary sentence completely contradicts the information in the reference document, requiring significant revision or removal to align with the reference document, the summary sentence is factually inconsistent with the reference document.
    \end{itemize}
    \vspace{-0.2cm}
    </errors>  \\ [3pt] \\ [3pt]
    <instructions>  \\ [3pt]
    Here is the instructions for writing your arguments: \\ [3pt]
    Follow these steps carefully to ensure a structured and thorough evaluation under your assigned role.  \\ [3pt]
    
    1. Read the reference document and summary sentence under your role:  \\ [3pt]
    \vspace{-0.5cm}
    \begin{itemize}
        \item[--] Carefully read the reference document and try to fully understand it. 
        \vspace{-0.2cm}
        \item[--] Compare each summary sentence to the reference document to identify evidence supporting its factual consistency. 
        \item[--] Refer to the list of errors in <errors></errors> tags to find evidence to support your claims(unfaithfulness), paying special attention to the listed error types and acceptable variations.
        \vspace{-0.2cm}
    \end{itemize}
    2. As a SKEPTIC, focus on identifying discrepancies: \\ [3pt]
    \vspace{-0.3cm}
    \begin{itemize}
        \item[--] Explicitly identify numbered sentences in the source document that contradict or fail to align with the summary sentence. 
        \vspace{-0.2cm}
        \item[--] Even if a perfect match cannot be found, select the closest sentence(s) that contain key elements (entities, relationships, events, quantities, or cause-effect relationships).
        \vspace{-0.2cm}
        \item[--] Always select at least one numbered sentence from the source document that highlights inconsistencies or raises doubts about the summary. 
    \end{itemize}
    \vspace{-0.2cm}    
    3. Provide a detailed explanation of your arguments:  \\ [5pt]
    \vspace{-0.5cm}
    \begin{itemize}
        \item[--]  For each summary sentence:
        \vspace{-0.2cm}
        \begin{itemize}
            \item[$\bullet$] Cite one or more numbered sentences from the reference document, even if only partial alignment exists. 
            \item[$\bullet$] Use the format "reference\_sentence\_number": [number1, number2] to explicitly indicate the reference of support.
            \item[$\bullet$] Include a "reason" that explicitly explains how the cited sentences align with the summary, focusing on entities,  relationships, and key details.
        \end{itemize}
        \vspace{-0.2cm}
        \item[--] Your explanation must:
        \vspace{-0.2cm}
        \begin{itemize}
            \item[$\bullet$] Be concise.
            \item[$\bullet$] Reference specific elements, such as: 
            \begin{itemize}
                \item Key details: Validate quantities, events, and cause-effect relationships. 
                \item Rephrasing accuracy: Argue that the summary retains the original meaning despite rephrasing.
                \item Word choice: Ensure terms like "only," "significant," or "most" match the intensity or scale of the reference document.
            \end{itemize}
        \end{itemize}
    \end{itemize}
    </instructions>
    
    Provide your answers in JSON format as shown below:  \\ [1pt]
    [\{ "summary\_sentence\_num": "0", "label": "1 ( if faithfulness 1, else 0 )", "error\_type": "no error ( if faithfulness no error )",  \\ [1pt]
        "reference\_sentence": ["sentence from reference document", "another sentence from reference document"], \\ [1pt]
        "reference\_sentence\_number": [0, 1], \\ [1pt]
        "reason": "The summary mentions the entity 'X', but the reference document refers to 'Y' [0]. Additionally, the relationship between 'A' and 'B' is misrepresented, as the reference indicates 'A causes B', not 'A results from B' [1]."\}, \\ [1pt]
      \{"summary\_sentence\_num": "N", "label": "1 ( if faithfulness 1, else 0 )",         "error\_type": "no error ( if faithfulness no error )",  \\ [1pt]
        "reference\_sentence": ["closest available sentence"], \\ [1pt]
        "reference\_sentence\_number": [10], \\ [1pt]
        "reason": "The summary states 'Event Z happened', but no reference to 'Event Z' is found in the reference [10]." \}] \\ [1pt] \\ [1pt]
    
    Reference document, divided into numbered sentences:  \\ [3pt]
    <document>  \\ [3pt]
    {input\_text} \\ [3pt]
    </document> \\ [3pt] \\ [3pt]
    
    Summary with {N} sentences: \\ [3pt]
    <summary> \\ [3pt]
    {summaries} \\ [3pt]
    </summary> \\ [3pt]
    \bottomrule
    \end{tabular}
    }
    \caption{Fact verification SKEPTIC prompt.}
    \label{tab:prompt_fv_skeptic}
\end{table*}
\begin{table*}[h]
    \scriptsize
    \centering
   \resizebox{\textwidth}{!}{%
   \begin{tabular}{p{0.9\linewidth}} 
    \toprule
    You are the ADJUDICATOR, an agent tasked with providing the final decision for the faithfulness of the summary by assessing the claims presented by the ADVOCATE and SKEPTIC. You are given the reference document provided in <document></document> tags, summary sentences in <summary></summary> tags and the opposing claims in <claim></claim> tags. Now, follow the instructions in <instructions></instructions> tags and"Make sure to always strive to deeply understand and remember the guidelines in <note></note> tags. Think critically and articulate your final decision. \\  [2pt] \\  [2pt]
        \vspace{-0.5cm}
    <note>
        \vspace{-0.2cm}
    \begin{enumerate}
        \item Faithfulness measures how accurately a summary sentence reflects the source document's information and content.
        \vspace{-0.2cm}
        \item The summary sentence should not have to use exact wording in the reference document as long as the original meaning is preserved.
        \vspace{-0.2cm}
        \item The summary sentence can paraphrase and use alternative expressions with preserving the original meaning.
        \vspace{-0.2cm}
        \item The summary sentence is factually consistent even if it omits specific details-one, some or all from reference document.
        \vspace{-0.2cm}
        \item The summary sentence is factually consistent even if it omits specific details-one, some or all from reference document.
        \vspace{-0.2cm}
        \item The summary sentence is factually consistent even if it modifies the level of specificity (using broader terms instead of detail and specific information, or more specific terms instead of the broader terms), maintaining the original information.
        \vspace{-0.2cm}
        \item The summary sentence is factually consistent even if it combines multiple pieces of information from different parts of the text maintaining the original meaning without contradiction.
        \vspace{-0.2cm}
        \item Even if the summary sentence draws reasonable implications, logical conclusions, or appropriate generalizations, it remains factually consistent with the reference document as long as these are explicitly supported by the original meaning.
        \vspace{-0.2cm}
    \end{enumerate}
    </note> \\  [2pt]
    
    <instructions>
    \vspace{-0.2cm}
    \begin{enumerate}
        \item Read the reference document and summary: 
        \vspace{-0.2cm}
         \begin{itemize}
            \item[--] Carefully review the reference document and the summary sentence provided.
            \item[--] Develop a comprehensive understanding of both the reference document and summary sentence and how it's been summarized.
        \end{itemize}
         \vspace{-0.2cm}
        \item Evaluate the validity of agent arguments:
         \vspace{-0.2cm}
         \begin{itemize}
            \item[--] Compare both agents' reasoning critically.
            \item[--] Validate the claims align with the reference document and avoid unsupported speculation.
            \item[--] Ensure the claims follow the guidelines in <note></note> tags.
        \end{itemize}
         \vspace{-0.2cm}
    \item Finalize your own judgment of the summary sentences. 
        \vspace{-0.2cm}
    \begin{itemize}
        \item[--]  Make a final decision on whether the summary sentence is factually consistent with the reference document, based on your understanding of reference document, summary sentence and the validation of the two opposing claims.
        \vspace{-0.2cm}
    \end{itemize}
    \item Provide your final decision as error type and label
    \vspace{-0.2cm}    
    \begin{itemize}
        \item[--] Assign an error type and label as follows:
        \item[--] error\_type: refer to the error\_type listed below:
        \begin{itemize}
            \item[$\bullet$] no error: no error found, and the summary is factually consistent with the reference document.
            \item[$\bullet$] out-of-article error: If the summary sentence introduces facts, subjective opinions, or biases that cannot be verified or confirmed by the reference document, the summary is factually inconsistent with the reference document.
            \item[$\bullet$]  entity error: If the summary sentence includes incorrect or misrepresented entities, such as names, numbers, or main subjects, the summary is factually inconsistent with the reference document.
            \item[$\bullet$] relation error: If the summary sentence contains incorrect semantic relationships, such as the use of wrong verbs, prepositions, or adjectives, which distort the relationships between entities, the summary sentence is factually inconsistent with the reference document.
            \item[$\bullet$] sentence error: If the summary sentence completely contradicts the information in the reference document, requiring significant revision or removal to align with the reference document, the summary sentence is factually inconsistent with the reference document.
        \end{itemize}
        \item[--] label:
        \begin{itemize}
            \item[$\bullet$] 1: faithfulness, assigned if the summary sentence has no error.
            \item[$\bullet$] 0: unfaithfulness, assigned if the summary contains any error (out-of-article, entity, relation, or sentence errors).
        \end{itemize}
        \item[--] In one or two sentences, explain why you agree with one agent's argument and disagree with the opposing agent's argument.
            \vspace{-0.2cm}
\end{itemize}
      \end{enumerate}
    
      </instructions> \\ [1pt]
    Provide your answers in JSON format as shown below:  \\ [1pt]
    [\{ "summary\_sentence\_num": "0", "label": "1 ( if faithfulness 1, else 0 )", "error\_type": "no error ( if faithfulness no error )",  \\ [1pt]
        "reason": "write reason of your decision briefly and concisely"\}, \\ [1pt]
      \{"summary\_sentence\_num": "N", "label": "1 ( if faithfulness 1, else 0 )",         "error\_type": "no error ( if faithfulness no error )",  \\ [1pt]
        "reason": "write reason of your decision briefly and concisely" \}] \\ [1pt] \\ [1pt]
    
    Reference document, divided into numbered sentences:  \\ [1pt]
    <document>  \\ [1pt]
    \{input\_text\} \\[1pt]
    </document> \\[1pt]
    Summary with {N} sentences: \\ [1pt]
    <summary> \\ [1pt]
    \{summaries\} \\[1pt]
    </summary> \\ [1pt]
    
    Here are the claims provided by ADVOCATE and SKEPTIC on for each summary sentence: \\ [1pt]
    <claim> \\ [1pt]
    \{claim\} \\ [1pt]
    </claim> \\ [1pt]
    \bottomrule
    \end{tabular}
    }
    \caption{Fact verification ADJUDICATOR prompt.}
    \label{tab:prompt_fv_adjudicator}
\end{table*}

\section{Faithfulness Error Types}
\label{sec:appendix_error_type}
We use error types from \,\citet{lee2024unisumeval}:

\smallskip
\noindent$\bullet$ \textbf{Out-of-article Error}: Introduces unverifiable facts, subjective opinions, or biases not supported by the source document.

\smallskip
\noindent$\bullet$ \textbf{Entity Error}: Misrepresents or includes incorrect entities, such as numbers or main subjects.

\smallskip
\noindent$\bullet$ \textbf{Relation Error}: Distorts the intended semantic relationships through incorrect use of verbs, prepositions, or adjectives.

\smallskip
\noindent$\bullet$ \textbf{Sentence Error}: Contradicts the source document, requiring major revisions or removal for factual alignment.

\section{Key-Fact Quality Assessment}
\label{sec:appendix_key-facts_quality_assessment}
Table~\ref{tab:key-fact_comparsion} illustrates a quantitative comparison between our domain-specific extraction methodology with the domain-agnostic approach from \citet{lee2024unisumeval}. The results indicate that our domain-specific extraction methodology encompasses a 32.61\% more in domain-relevant key-fact identification relative to the baseline. Notably, the domain-specific strategy systematically excludes 16.83\% of key-facts identified by the domain-agnostic method, as these elements lie beyond the scope of domain-specific criteria. This selective filtering highlights our methodology's precision in capturing domain-relevant information.

Additionally, we qualitatively assess the effectiveness of our domain-specific methodology compared to the domain-agnostic baseline via A/B preference testing with two expert annotators. We provide them with the English source document and two sets of key-facts: one extracted by the domain-agnostic baseline and the other identified by our domain-specific methodology. To minimize presentation bias, we present the two sets in randomized order, and the annotators are blind to the source of each set.
They independently judge which set is more useful for summarization and better captures the essential content of the source document. To assess the consistency of these subjective judgments, we compute IAA using Cohen’s kappa. Table~\ref{tab:key-fact_preference} presents the comprehensive results of the A/B preference test and Cohen's kappa.
Across all domains, \algname{} is consistently preferred by annotators, with preference rates ranging from 62\% to 90.0\%. Notably, Medical Literature shows a higher preference ratio (90.0\%) compared to Meeting (62.0\%), demonstrating the necessity of domain-specialized extraction in fields where domain expertise significantly influences content understanding.


\section{Benchmark Evaluation Formula}
\label{sec:appendix_formula}
\subsection{Evaluation Metrics for Summarization}
We evaluate performance using five key metrics: faithfulness, completeness, conciseness, domain stability, and language stability. The first three metrics follow the methodology established by \,\citet{song2024finesure}. 

\paragraph{Faithfulness} We consider a source document $D$, and its generated summary $S$ containing $N$ sentences $\{s_1, s_2, ..., s_N\}$. Based on human annotation, we identify $S_{Fact} \subseteq S$, which is the subset of $S$ containing only factually accurate sentences. The faithfulness score is calculated as the ratio of factually correct sentences to total sentences:
\[
    \text{Faithfulness}(D,S) = \frac{|S_{Fact}|}{|S|} 
\]
\paragraph{Completeness and Conciseness} Let $K$ be the set of key-facts $\{k_1, k_2, ..., k_M\}$ from the source document. Based on the results of key-fact alignment, we can define bipartite graph $G = (K, S, E)$, where $E$ represents edges between key-facts $K$ and summary sentences $S$, $\{(k,s): k\rightarrow s | k \in K \wedge s \in S\}$ with $k \rightarrow s$ signifying that key-fact $k$ is entailed in $s$. 

The completeness score measures how many key facts are captured in the summary:
\[
    \text{Completeness}(K,S) = \frac{|\{k|(k,s) \in E\}|}{|K|} 
\]
The conciseness score measures how efficiently the summary sentences convey key facts:
\[
    \text{Conciseness}(K,S) = \frac{|\{s|(k,s) \in E\}|}{|S|} 
\]

\paragraph{Domain Stability} 
We introduce an improved approach to measure domain stability score, addressing limitations in \,\citet{lee2024unisumeval}. While the previous approach only considered the gap between the highest and lowest scores, our method accounts for overall performance variability.
To achieve this, we use the coefficient of variation (CV), which provides a normalized measure of dispersion by comparing the standard deviation to the mean. 
We calculate CV for three evaluation dimensions respectively across domains as follows below formulation.

For a given performance score across domains $S_E = \{S_{E,1}, S_{E,2}, ..., S_{E,d}\}$, where $d = 1, ... 6$ denotes the domain index and $E$ refers faithfulness, completeness, and conciseness. We calculate the Instability score for each evaluation dimension $S_E$:
\[
    \text{Instability}(S_E) = \frac{\sigma_{S_E}}{\mu_{S_E}} \times 100,
\]
where $\sigma_{S_E}$ is the standard deviation and $\mu_{S_E}$ is the mean of the scores for dimension $E$ across domain.
To formulate Domain Stability, we rescaled the Instability($S_{E}$) to ensure the stability measure remains within a meaningful range. Specifically, we define \textit{Domain Stability} as:
\[
    \text{Domain Stability}(S_{E}) = \frac{100}{1 + \text{Instability}(S_{E})}
\]
For the composite score, we average the Domain Stability of the three evaluation dimensions.

\paragraph{Language Stability}
Language stability evaluates how consistently a model performs across different languages. We apply the same mathematical framework as Domain Stability, using the coefficient of variation (CV) and rescaling approach. 
For each evaluation dimension $E$, we first compute $S_{E,L}$, the average score across domains for each language $L$:
\[
S_{E,L} = \frac{1}{D} \sum_{d=1}^{D} S_{E,L,d}, 
\]
where $S_{E,L,d}$ represents the score for evaluation dimension $E$ in language $L$ and domain $d$, and $D$ is the total number of domains.
Then we calculate the instability score with CV, which captures the performance fluctuation across different languages:
\[
    \text{Instability}(S_E) = \frac{\sigma_{S_E}}{\mu_{S_E}} \times 100,
\]
where $\sigma_{S_E}$ is the standard deviation and $\mu_{S_E}$ is the mean of the scores for dimension $E$ across languages. 

Similar to Domain Stability, we transform the Instability score to ensure that the stability measure remains within a meaningful range:
\[
    \text{Language Stability}(S_{E}) = \frac{100}{1 + \text{Instability}(S_{E})}
\]
The composite score is calculated by averaging the Language Stability across three evaluation dimensions. Similar to domain stability, it reflects the model's performance variations across different languages.

\subsection{Evaluation Metrics for Automatic Evaluator}
\paragraph{Pearson Correlation}
\label{sec:appendix_pearson}
We compute summary-level correlations using the Pearson correlation coefficient to assess the alignment between automated and human evaluation, following recent work\,\cite{liu2023geval, song2024finesure,lee2024unisumeval}. For each summary $i$, we analyze the correlation between automated evaluation scores ($x_i$) and human evaluation scores ($y_i$). The summary-level correlation $\rho$ is calculated as:
\[
    \rho = Cor([x_1, x_2, ..., x_n], [y_1, y_2, ..., y_n])
\]
where $Cor$ represents the Pearson correlation coefficient, where $n$ is the total number of summaries.

\section{Self-Evaluation Bias Detail}
\label{sec:appendix_self-preference_bias}
To investigate self-evaluation bias in summary evaluation, we employ four large language models (GPT-4o, Claude-3.5-Sonnet, Llama-3.1-70B, and Qwen-2.5-72B) as both evaluators and summarizers. We measure self-evaluation bias by comparing how each model rates its own summaries versus those generated by other models.

First, we calculate composite scores by averaging three evaluation dimensions (faithfulness, completeness, and conciseness). Second, each model’s self-evaluation score is obtained by averaging the composite scores of its own summaries across different domains. Similarly, peer-evaluation scores are computed by averaging the composite scores given by other models. 
Third, self-evaluation bias is then determined by subtracting the peer-evaluation scores from the self-evaluation scores. 
To assess statistical significance, we conduct the t-test comparing self-evaluation and peer-evaluation scores. Biases that are statistically significant (p < 0.05) are highlighted in the visualizations in Figure~\ref{fig:self-preference}.

\section{Additional Analysis}

\subsection{Human Annotation Result Detail}
\label{sec:appendix_human_annotation_result_detail}
\begin{table*}[]
\scriptsize
\renewcommand{\arraystretch}{1.2}
\setlength{\tabcolsep}{2pt}
\centering
\resizebox{\textwidth}{!}{%
\begin{tabular}{ccccccccccccccccc}
\toprule
\multirow{2}{*}{Model Type} &
  \multirow{2}{*}{Summarizer} &
  \multicolumn{7}{c}{English} &
  \multicolumn{7}{c}{Chinese} &
  \multicolumn{1}{l}{\multirow{2}{*}{\begin{tabular}[c]{@{}l@{}}Language \\ Stability\end{tabular}}} \\
  \cmidrule(lr){3-9} \cmidrule(lr){10-16}
 &
   &
  News &
  Med Lit &
  Report &
  Booking &
  Meeting &
  Interview &
  Domain* &
  News &
  Med Lit &
  Report &
  Booking &
  Meeting &
  Interview &
  Domain* &
  \\
  \midrule

\multirow{4}{*}{\begin{tabular}[c]{@{}c@{}}Prop-\\ rietary \\ LLMs\end{tabular}} &
  GPT-4o &
  86.30 &
  80.10 &
  82.50 &
  95.10 &
  79.90 &
  92.70 &
  92.97 &
  87.10 &
  78.01 &
  80.00 &
  88.20 &
  56.70 &
  80.50 &
  87.31 &
  93.81 \\
 &
  Claude 3.5 \textsubscript{Sonnet} &
  \textbf{96.20} &
  83.70 &
  89.50 &
  94.40 &
  \textbf{80.60} &
  92.10 &
  \textbf{93.58} &
  \textbf{88.90} &
  \textbf{86.27} &
  73.20 &
  84.00 &
  \textbf{75.40} &
  \textbf{88.30} &
  \textbf{92.45} &
  94.75\\
 &
  Gemini 1.5  \textsubscript{Pro} &
  85.60 &
  \textbf{86.90 }&
  88.30 &
  \textbf{97.40} &
  74.70 &
  84.20 &
  92.18 &
  87.00 &
  84.64 &
  82.80 &
  \textbf{92.00} &
  60.60 &
  74.50 &
  87.74 &
  \textbf{95.21} \\
  \cmidrule(lr){2-17}
  & Average & 89.37 & 83.57 & 86.77 & 95.63 & 78.40 & 89.67 & 92.91 & 87.67 & 82.97 & 78.67 & 88.07 & 64.23 & 81.10 & 89.17 & 94.59\\
  \midrule
  
\multirow{4}{*}{\begin{tabular}[c]{@{}c@{}}Open\\ source \\ LLMs\end{tabular}} &
  Gemma 2  \textsubscript{27B} &
  94.90 &
  84.10 &
  84.00 &
  93.60 &
  72.20 &
  93.30 &
  90.87 &
  87.00 &
  81.30 &
  72.90 &
  77.20 &
  66.80 &
  86.00 &
  90.94 &
  93.24 \\
 &
  Llama 3.1  \textsubscript{70B}&
  87.70 &
  76.40 &
  \textbf{92.90} &
  89.30 &
  78.50 &
  76.20 &
  91.93 &
  84.90 &
  77.70 &
  72.00 &
  79.00 &
  67.10 &
  81.90 &
  92.18 &
  94.66 \\
 &
  Qwen 2.5  \textsubscript{72B} &
  95.10 &
  81.30 &
  91.40 &
  91.50 &
  73.70 &
  \textbf{94.70} &
  91.11 &
  82.40 &
  86.07 &
  86.00 &
  80.20 &
  62.50 &
  84.60 &
  89.92 &
  93.95 \\
  \cmidrule(lr){2-17}
  &Average  & 92.57 & 80.60 & 89.43 & 91.47 & 74.80 & 88.07 & 91.30 & 84.77 & 81.69 & 76.97 & 78.80 & 65.47 & 84.17 & 91.01 & 93.95\\
  \midrule
\multirow{3}{*}{\begin{tabular}[c]{@{}c@{}}Non\\ LLMs\end{tabular}} &
  mT5 &
  22.00 &
  12.00 &
  2.00 &
  34.00 &
  4.00 &
  0.00 &
  48.02 &
  68.00 &
  18.00 &
  32.00 &
  2.00 &
  14.00 &
  18.00 &
  52.41 &
  67.20 \\
 &
  BART &
  93.90 &
  76.30 &
  83.00 &
  76.00 &
  50.30 &
  85.90 &
  83.87 &
  - &
  - &
  - &
  - &
  - &
  - &
  - &
  -\\
  \cmidrule(lr){2-17}
  & Average & 57.95 & 44.15 & 42.50 & 55.00 & 27.15 & 42.95 & 65.94 & - & -  & -  &  - & -  & -  & -  &\\ 
  \bottomrule
\end{tabular}
}
\caption{Faithfulness score across six domains and two languages. Domain*: domain stability}
\label{tab:appendix_faithfulness}
\vspace*{0.5cm}
\end{table*}

\begin{table*}[]
\scriptsize
\renewcommand{\arraystretch}{1.2}
\setlength{\tabcolsep}{2pt}
\centering
\resizebox{\textwidth}{!}{%
\begin{tabular}{ccccccccccccccccc}
\toprule
\multirow{2}{*}{Model Type} &
  \multirow{2}{*}{Summarizer} &
  \multicolumn{7}{c}{English} &
  \multicolumn{7}{c}{Chinese} &
  \multicolumn{1}{l}{\multirow{2}{*}{\begin{tabular}[c]{@{}l@{}}Language \\ Stability\end{tabular}}} \\
    \cmidrule(lr){3-9} \cmidrule(lr){10-16}
 &
   &
  News &
  Med Lit &
  Report &
  Booking &
  Meeting &
  Interview &
  Domain* &
  News &
  Med Lit &
  Report &
  Booking &
  Meeting &
  Interview &
  Domain* &
   \\
  \midrule
\multirow{4}{*}{\begin{tabular}[c]{@{}c@{}}Prop-\\ rietary \\ LLMs\end{tabular}} &
  GPT-4o &
  55.73 &
  \textbf{41.22} &
  40.25 &
  68.19 &
  50.10 &
  44.65 &
  \textbf{82.48} &
  55.80 &
  32.63 &
  33.83 &
  47.58 &
  39.38 &
  37.92 &
  82.24 &
  87.95 \\
 &
  Claude 3.5 \textsubscript{Sonnet} &
  \textbf{65.49} &
  39.96 &
  42.88 &
  68.77 &
  48.40 &
  50.51 &
  81.61 &
  55.89 &
  \textbf{41.06} &
  35.20 &
  71.43 &
  46.87 &
  47.17 &
  79.58 &
  95.93 \\
 &
  Gemini 1.5 \textsubscript{Pro} &
  64.43 &
  39.92 &
  \textbf{44.43} &
  \textbf{78.35} &
  48.40 &
  \textbf{53.30} &
  79.33 &
  \textbf{57.74} &
  38.87 &
  \textbf{48.49} &
  \textbf{72.20} &
  \textbf{51.43} &
  \textbf{52.54} &
  \textbf{82.88} &
  \textbf{98.38} \\
  \cmidrule(lr){2-17}
  & Average & 61.88 & 40.37 & 42.52 & 71.77 & 48.97 & 49.49 & 81.14 & 56.48 & 37.52 & 39.17 & 63.74 & 45.89 & 45.88 & 81.57 & 94.09 \\
  \midrule
\multirow{4}{*}{\begin{tabular}[c]{@{}c@{}}Open\\ source \\ LLMs\end{tabular}} &
  Gemma 2 \textsubscript{27B} &
  52.26 &
  23.67 &
  25.86 &
  56.95 &
  35.83 &
  27.77 &
  72.19 &
  43.23 &
  23.65 &
  21.43 &
  51.25 &
  23.72 &
  29.77 &
  72.41 &
  90.93 \\
 &
  Llama 3.1 \textsubscript{70B} &
  51.96 &
  29.34 &
  35.91 &
  51.66 &
  29.48 &
  36.76 &
  79.25 &
  24.59 &
  21.71 &
  21.26 &
  35.06 &
  20.46 &
  25.60 &
  82.05 &
  75.84 \\
 &
  Qwen 2.5 \textsubscript{72B} &
  55.81 &
  37.73 &
  39.58 &
  70.12 &
  \textbf{55.24} &
  46.74 &
  80.79 &
  47.69 &
  34.13 &
  33.17 &
  61.48 &
  34.35 &
  34.84 &
  78.15 &
  86.74 \\
  \cmidrule(lr){2-17}
  & Average & 53.34 & 30.25 & 33.78 & 59.58 & 40.18 & 37.09 & 77.41 & 38.50 & 26.50 & 25.29 & 49.26 & 26.18 & 30.07 & 77.54 & 84.50\\
  \midrule
\multirow{3}{*}{\begin{tabular}[c]{@{}c@{}}Non\\ LLMs\end{tabular}} &
  mT5 &
  9.66 &
  0.67 &
  1.68 &
  6.09 &
  3.07 &
  1.56 &
  52.36 &
  9.97 &
  0.00 &
  1.67 &
  0.97 &
  2.12 &
  2.43 &
  44.34 &
  83.51 \\
 &
  BART &
  32.97 &
  14.93 &
  13.15 &
  45.49 &
  22.74 &
  20.79 &
  67.16 &
  - &
  - &
  - &
  - &
  - &
  - &
  - &
  - \\
  \cmidrule(lr){2-17}
  & Average & 21.32 & 7.80 & 7.42 & 25.79 & 12.91 & 11.18 & 59.76 & -  & -  &  - &  - & -  & -  & -  & \\
  \bottomrule
\end{tabular}
}
\caption{Completeness score across six domains and two languages. Domain*: domain stability}
\label{tab:appendix_completeness}
\vspace*{0.5cm}
\end{table*}

\begin{table*}[]
\scriptsize
\renewcommand{\arraystretch}{1.2}
\setlength{\tabcolsep}{2pt}
\centering
\resizebox{\textwidth}{!}{%
\begin{tabular}{ccccccccccccccccc}
\toprule
\multirow{2}{*}{Model Type} &
  \multirow{2}{*}{Summarizer} &
  \multicolumn{7}{c}{English} &
  \multicolumn{7}{c}{Chinese} &
  \multicolumn{1}{l}{\multirow{2}{*}{\begin{tabular}[c]{@{}l@{}}Language \\ Stability\end{tabular}}} \\
    \cmidrule(lr){3-9} \cmidrule(lr){10-16}
 &
   &
  News &
  Med Lit &
  Report &
  Booking &
  Meeting &
  Interview &
  Domain* &
  News &
  Med Lit &
  Report &
  Booking &
  Meeting &
  Interview &
  Domain* &
   \\
  \midrule
\multirow{4}{*}{\begin{tabular}[c]{@{}c@{}}Prop-\\ rietary \\ LLMs\end{tabular}} &
  GPT-4o &
  85.73 &
  83.40 &
  81.20 &
  74.76 &
  73.51 &
  69.27 &
  92.39 &
  84.20 &
  \textbf{79.85} &
  77.75 &
  74.16 &
  65.00 &
  68.91 &
  91.33 &
  97.30 \\
 &
  Claude 3.5 \textsubscript{Sonnet} &
  90.73 &
  74.34 &
  80.59 &
  87.26 &
  72.18 &
  75.43 &
  91.42 &
  84.72 &
  79.05 &
  73.56 &
  \textbf{89.33} &
  66.71 &
  75.05 &
  90.57 &
  \textbf{98.23} \\
 &
  Gemini 1.5 \textsubscript{Pro} &
  88.59 &
  83.18 &
  79.71 &
  \textbf{89.90} &
  69.88 &
  \textbf{80.01} &
  91.86 &
  88.07 &
  74.12 &
  77.71 &
  87.35 &
  \textbf{71.35} &
  \textbf{77.98} &
  \textbf{92.04} &
  97.90 \\
  \cmidrule(lr){2-17}
  & Average & 88.35 & 80.31 & 80.50 & 83.97 & 71.86 & 74.90 & 91.89 & 85.66 & 77.67 & 76.34 & 83.61 & 67.69 & 73.98 & 91.31 & 97.81 \\
  \midrule
\multirow{4}{*}{\begin{tabular}[c]{@{}c@{}}Open\\ source \\ LLMs\end{tabular}} &
  Gemma 2 \textsubscript{27B} &
  88.07 &
  70.69 &
  72.47 &
  75.13 &
  69.40 &
  66.23 &
  90.58 &
  81.47 &
  71.46 &
  59.35 &
  82.00 &
  52.47 &
  71.36 &
  85.49 &
  96.22 \\
 &
  Llama 3.1 \textsubscript{70B} &
  \textbf{93.03} &
  79.33 &
  86.05 &
  86.20 &
  67.87 &
  75.35 &
  90.05 &
  85.07 &
  73.72 &
  72.22 &
  85.67 &
  66.35 &
  67.73 &
  89.95 &
  94.71 \\
 &
  Qwen 2.5 \textsubscript{72B} &
  91.63 &
  \textbf{84.38} &
  \textbf{87.29} &
  87.67 &
  \textbf{83.02} &
  75.20 &
  \textbf{93.81} &
  \textbf{91.53} &
  78.06 &
  \textbf{86.53} &
  83.33 &
  66.86 &
  70.73 &
  89.37 &
  95.59 \\
  \cmidrule(lr){2-17}
  & Average & 90.91 & 78.13 & 81.94 & 83.00 & 73.43 & 72.26 & 91.48 & 86.02 & 74.41 & 72.70 & 83.67 & 61.89 & 69.94 & 88.27 & 95.51 \\
  \midrule
\multirow{3}{*}{\begin{tabular}[c]{@{}c@{}}Non\\ LLMs\end{tabular}} &
  mT5 &
  46.00 &
  8.00 &
  28.00 &
  42.00 &
  20.00 &
  28.00 &
  67.17 &
  80.00 &
  0.00 &
  36.00 &
  12.00 &
  18.00 &
  20.00 &
  49.54 &
  97.55 \\
 &
  BART &
  89.00 &
  81.00 &
  86.00 &
  88.00 &
  60.00 &
  66.90 &
  86.58 &
  - &
  - &
  - &
  - &
  - &
  - &
  - &
  -\\
  \cmidrule(lr){2-17}
  & Average & 67.50 & 44.50 & 57.00 & 65.00 & 40.00 & 47.45 & 76.88 & - & - &  -&  -& - & - & - &  - \\
  \bottomrule
\end{tabular}
}
\caption{Conciseness score across six domains and two languages. Domain*: domain stability}
\label{tab:appendix_conciseness}
\vspace*{0.5cm}
\end{table*}

We provide the eight summarizer performances across all six domains and languages of faithfulness, completeness, and conciseness in Tables~\ref{tab:appendix_faithfulness} - ~\ref{tab:appendix_conciseness}. We
present additional domain-level findings. 
\paragraph{Faithfulness Score}
Table~\ref{tab:appendix_faithfulness} shows proprietary LLMs outperform open-source and non-LLMs across domains and languages.
Specifically, Claude-3.5-Sonnet achieves optimal performance stability among proprietary LLMs across all domains and languages.
A marginal decline in faithfulness is observed in News, Booking and Meeting domains, while an increase is noted in Report and Interview domains from English to Chinese. Therefore, further research efforts should target both language-specific enhancements and domain-language adaptation strategies.

\paragraph{Completeness Score}
Table~\ref{tab:appendix_completeness} shows proprietary LLMs' superior performance across all domains and languages, with Gemini-1.5-Pro excelling in Report, Booking, and Interview domains. 
However, a decline in completeness scores is observed across all domains and models from English to Chinese, with varying degrees of intensity depending on the domain. 
The observed decline in completeness scores indicates challenges in retaining essential information in Chinese, emphasizing the need for domain-aware language improvements to enhance performance.

\paragraph{Conciseness Score}
Table~\ref{tab:appendix_conciseness} demonstrates proprietary LLMs achieve higher conciseness scores compared to open-source and non-LLMs across domains. 
While the overall decline in conciseness from English to Chinese is minimal, the Meeting domain experiences a significant drop, with open-source LLMs performing the worst in this domain compared to others. This highlights the need for domain-specific refinements to improve conciseness in open-source LLMs.


\subsection{Comparison with QA- and NLI-based Auto-evaluator}
\begin{table*}[hbt!]
\scriptsize
\centering
\setlength{\tabcolsep}{11pt}
\begin{tabular}{ccccccccc}
\toprule
Model Type & Evaluator & News & Medical Lit. & Report & Booking & Meeting & Interview & Domain Stability \\
\midrule
QA-Based                   & QAFactEval & 0.53 & 0.53 & 0.37 & 0.36 & 0.33 & 0.44 & 83.08 \\
\midrule

\multirow{2}{*}{NLI-Based} & AlignScore & \textbf{0.73} & 0.73 & 0.63 & 0.28 & 0.56 & 0.51 & 77.58 \\
                           & MiniCheck  & 0.59 & \textbf{0.75} & 0.56 & 0.36 & 0.41 & 0.42 & 77.72 \\
\midrule
                           
\multirow{2}{*}{LLM-Based} & G-Eval     & 0.52 & 0.67 & 0.67 & 0.37 & 0.38 & 0.46 & 79.10 \\
                           & FineSurE   & 0.42 & 0.62 & \textbf{0.69} & \textbf{0.48} & \textbf{0.70} &\textbf{ 0.60 }& \textbf{83.75}\\
\bottomrule
\end{tabular}

\caption{Correlation between automatic evaluator and human evaluation.}
\label{tab:appendix_auto-eval_en}
\end{table*}
In Table~\ref{tab:appendix_auto-eval_en}, we provide the performance of automated evaluators, including QA-based (QAFactEval\,\cite{fabbri2022qafacteval}), NLI-based (AlignScore\,\cite{zha2023alignscore}, MiniCheck\,\cite{tang2024minicheck}), and LLM-based (G-Eval\,\cite{liu2023geval}, FineSurE\,\cite{song2024finesure}) with implementations based on GPT-4o.

The QA-based and NLI-based evaluators specialize in measuring faithfulness but have limitations in assessing completeness and conciseness. Therefore, we focus on comparing their faithfulness scores across six domains. We compare the scores obtained from automatic evaluation with human-annotated labels in \algname{}. Specifically, we report Pearson correlation values for the summary-level faithfulness percentage scores. Due to the language limitations of these evaluators, we only provide results for English summaries. Among all methods, FineSurE achieves the highest domain stability (83.75) and demonstrates superior performance in four domains such as Report, Booking, Meeting, and Interview.

\subsection{Detailed Performance of LLM-Based Auto-Evaluator}
\label{sec:appendix_auto-eval_llm}
\begin{table*}[hbt!]
\footnotesize
\centering
\renewcommand{\arraystretch}{1.4}
\setlength{\tabcolsep}{2pt}
\resizebox{\textwidth}{!}{%
\begin{tabular}{cccccccccccccccccc}
\toprule
\multirow{2}{*}{Dimension} &
  \multirow{2}{*}{\begin{tabular}[c]{@{}c@{}}Model\\ Type\end{tabular}} &
  \multirow{2}{*}{\begin{tabular}[c]{@{}c@{}}Automatic\\ Evaluator\end{tabular}} &
  \multicolumn{7}{c}{English} &
  \multicolumn{7}{c}{Chinese} &
  \multirow{2}{*}{\begin{tabular}[c]{@{}c@{}}Language \\ Stability\end{tabular}} \\
  \cmidrule(lr){4-10} \cmidrule(lr){11-17} 
  
 &
   &
   &
  News &
  Medical Lit. &
  Report &
  Booking &
  Meeting &
  Interview &
  Domain* &
  News &
  Medical Lit. &
  Report &
  Booking &
  Meeting &
  Interview &
  Domain* &
   \\
   \midrule
\multirow{4}{*}{Faithfulness} &
  \multirow{2}{*}{\begin{tabular}[c]{@{}c@{}}Proprietary \\LLMs\end{tabular}} &
  G-Eval &
  0.62 &
  0.61 &
  0.62 &
  0.75 &
  \textbf{0.60} &
  0.66 &
  91.19 &
  0.26 &
  0.52 &
  0.44 &
  0.62 &
  0.43 &
  0.53 &
  78.29 &
  \textbf{94.17} \\
 &
   &
  FineSurE &
  \textbf{0.72} &
  0.64 &
  \textbf{0.85} &
  0.64 &
  0.58 &
  \textbf{0.78} &
  87.29 &
  \textbf{0.39} &
  \textbf{0.64} &
  \textbf{0.57} &
  \textbf{0.74} &
  \textbf{0.52} &
  \textbf{0.66} &
  \textbf{82.18} &
  88.71 \\
  \cmidrule(lr){2-18}
  
 &
  \multirow{2}{*}{\begin{tabular}[c]{@{}c@{}}Open-source \\ LLMs\end{tabular}} &
  G-Eval &
  0.69 &
  \textbf{0.65} &
  0.72 &
  \textbf{0.78} &
  0.56 &
  0.67 &
  \textbf{92.46 }&
  0.32 &
  0.52 &
  \textbf{0.57} &
  0.73 &
  0.42 &
  0.54 &
  76.33 &
  87.77 \\
 &
   &
  FineSurE &
  0.68 &
  0.54 &
  0.65 &
  0.63 &
  0.57 &
  0.75 &
  88.11 &
  0.37 &
  0.52 &
  0.54 &
  0.66 &
  0.44 &
  0.60 &
  81.80 &
  87.00 \\
  \midrule
\multirow{4}{*}{Completeness} &
  \multirow{2}{*}{\begin{tabular}[c]{@{}c@{}}Proprietary \\LLMs\end{tabular}} &
  G-Eval &
  0.62 &
  0.61 &
  0.62 &
  0.75 &
  0.60 &
  0.66 &
  90.56 &
  0.71 &
  0.40 &
  0.67 &
  0.66 &
  0.64 &
  0.65 &
  77.93 &
  89.75 \\
 &
   &
  FineSurE &
  \textbf{0.79} &
  \textbf{0.76} &
  \textbf{0.81 }&
  \textbf{0.89} &
 \textbf{ 0.83 }&
  \textbf{0.82 }&
  \textbf{94.79} &
  \textbf{0.83 }&
  \textbf{0.75} &
  \textbf{0.75} &
 \textbf{ 0.90} &
  \textbf{0.70 }&
 \textbf{ 0.83} &
  \textbf{90.56} &
  \textbf{95.27} \\
  \cmidrule(lr){2-18}
  
 &
  \multirow{2}{*}{\begin{tabular}[c]{@{}c@{}}Open-source \\ LLMs\end{tabular}} &
  G-Eval &
  0.69 &
  0.65 &
  0.72 &
  0.78 &
  0.56 &
  0.67 &
  92.25 &
  0.69 &
  0.46 &
  0.59 &
  0.75 &
  0.57 &
  0.59 &
  82.25 &
  92.19 \\
 &
   &
  FineSurE &
  0.66 &
  0.54 &
  0.61 &
  0.81 &
  0.72 &
  0.61 &
  89.29 &
  0.62 &
  0.19 &
  0.35 &
  0.81 &
  0.46 &
  0.65 &
  76.46 &
  88.15\\
  \midrule
\multirow{4}{*}{Completeness} &
  \multirow{2}{*}{\begin{tabular}[c]{@{}c@{}}Propprietary \\LLMs\end{tabular}} &
  G-Eval &
  0.64 &
  0.58 &
  0.48 &
  0.43 &
  0.28 &
  0.48 &
  85.20 &
  0.46 &
  0.59 &
  0.44 &
  0.47 &
  0.50 &
  0.58 &
  78.44 &
  85.40 \\
 &
   &
  FineSurE &
  \textbf{0.80 }&
  \textbf{0.72} &
  \textbf{0.57} &
  \textbf{0.76} &
  \textbf{0.68} &
  \textbf{0.72 }&
  \textbf{89.42 }&
  \textbf{0.78} &
  \textbf{0.74} &
  \textbf{0.64} &
  \textbf{0.84} &
  \textbf{0.60} &
  \textbf{0.74} &
  \textbf{88.62} &
  \textbf{97.13}\\
  \cmidrule(lr){2-18}
 &
  \multirow{2}{*}{\begin{tabular}[c]{@{}c@{}}Open-source \\ LLMs\end{tabular}} &
  G-Eval &
  0.65 &
  0.63 &
  0.51 &
  0.46 &
  0.51 &
  0.47 &
  86.33 &
  0.30 &
  0.65 &
  0.35 &
  0.72 &
  0.45 &
  0.54 &
  75.50 &
 89.80 \\
 &
   &
  FineSurE &
  0.74 &
  0.46 &
  0.46 &
  0.67 &
  0.56 &
  0.56 &
  84.25 &
  0.62 &
  0.19 &
  0.35 &
  0.81 &
  0.46 &
  0.65 &
  75.82 &
  92.65 \\
  \bottomrule
\end{tabular}%
}
\caption{Human alignment performance of automatic evaluators across three evaluation dimensions, six domains and two languages. Domain*: Domain Stability}
\label{tab:appendix_llm_domain}
\end{table*}
Table~\ref{tab:appendix_llm_domain} compares the performance of two LLM-based automatic evaluators, G-Eval and FineSurE, across two model types, six domains, and two languages. We report the average scores of GPT-4o and Claude-3.5-Sonnet for proprietary LLMs and Llama-3.1-70B and Qwen-2.5-72B for open-source LLMs. While FineSurE, implemented with proprietary LLMs, often demonstrates strong performance across three evaluation dimensions, its effectiveness varies across domains, particularly in faithfulness evaluation. In some cases, such as the Booking domain, G-Eval (using open-source LLMs) surpasses FineSurE by a significant margin. 
This highlights the inherent variability in LLM-based evaluation accuracy and reinforces that no single LLM or automatic evaluator consistently outperforms others across all dimensions, domains, and languages.

\subsection{Comparison with Similarity-based Metric}
\label{sec:appendix_auto-eval_trad}

\begin{table*}[]
\scriptsize
\setlength{\tabcolsep}{4pt}
\centering
\begin{tabular}{cccccccccccccc}
\toprule
\multirow{2}{*}{Dimension}  & \multirow{2}{*}{Evaluator} & \multicolumn{6}{c}{English}                              & \multicolumn{6}{c}{Chinese}                              \\
\cmidrule(lr){3-8} \cmidrule(lr){9-14}
       &                      & News  & Med Lit & Report & Booking & Meeting & Interview & News  & Med Lit & Report & Booking & Meeting & Interview \\
\midrule
\multirow{5}{*}{Faithfulness} & ROUGE-1               & -0.16 & 0.35    & 0.38   & -       & -0.1    & 0.06      & -0.07 & 0.12    & -0.09  & -       & 0.16    & 0.01      \\
 & ROUGE-2    & -0.05 & -0.08 & -0.1  & -    & 0.13  & -0.17 & 0.07  & 0.1   & 0.02  & -    & 0.03  & -0.35 \\
 & ROUGE-L    & -0.24 & -0.18 & 0.05  & -    & 0.02  & -0.18 & 0.04  & -0.06 & 0.18  & -    & 0.15  & -0.37 \\
 & BERTScore & -0.1  & -0.16 & 0.02  & -    & 0.11  & -0.15 & 0     & 0.05  & -0.06 & -    & 0.03  & -0.16 \\
 \cmidrule(lr){2-14}
 & FineSurE  & 0.74* & 0.68* & 0.52* & 0.75* & 0.69* & 0.64* & 0.42* & 0.60* & 0.62* & 0.70* & 0.48* & 0.69*  \\
\midrule
 
\multirow{5}{*}{Completeness}  & ROUGE-1               & 0.03  & 0       & 0      & -       & 0.36    & 0.16      & -0.23 & 0.2     & 0.33   & -       & 0.59*   & 0.37      \\
 & ROUGE-2    & -0.16 & 0.24  & 0.11  & -    & 0.53* & 0.24  & -0.3  & 0.22  & 0.28  & -    & 0.39  & 0.28  \\
 & ROUGE-L    & -0.1  & 0.04  & 0.07  & -    & 0.41* & 0.07  & -0.23 & 0.04  & 0.48* & -    & 0.41* & 0.14  \\
 & BERTScore & 0.05  & 0.13  & 0.06  & -    & 0.54* & 0.2   & -0.11 & 0.16  & 0.38  & -    & 0.39  & 0.29  \\
 \cmidrule(lr){2-14}
 & FineSurE  &0.79* & 0.75* & 0.81* & 0.90* & 0.85* & 0.80* & 0.78* & 0.66* & 0.64* & 0.90*& 0.66* & 0.75*  \\
\midrule
 
\multirow{5}{*}{Conciseness}  & ROUGE-1               & 0.51* & 0.54*   & 0.23*  & -       & 0.53*   & 0.55*     & 0.42* & 0.58*   & 0.4*   & -       & 0.66*   & 0.69*     \\
 & ROUGE-2    & 0.41* & 0.39* & -0.04 & -    & 0.44* & 0.61* & 0.22* & 0.45* & -0.02 & -    & 0.54* & 0.72* \\
 & ROUGE-L    & 0.29* & 0.41* & -0.01 & -    & 0.43* & 0.47* & 0.13  & 0.4*  & 0.05  & -    & 0.45* & 0.52* \\
 & BERTScore & 0.33* & 0.42* & 0.02  & -    & 0.43* & 0.6*  & 0.13  & 0.48* & -0.11 & -    & 0.54* & 0.65* \\
 \cmidrule(lr){2-14}
 & FineSurE  & 0.75* & 0.68* & 0.52* & 0.75* & 0.69* & 0.64* & 0.80* & 0.70* & 0.64* & 0.82* & 0.59* & 0.74* \\
 \bottomrule
\end{tabular}%
\caption{Correlation between similarity-based metric and human evaluation score. *: p-value \textless 0.05}
\label{tab:appendix_trad_auto-eval}
\end{table*}

We analyze the summary-level agreement between human scores and conventional similarity-based metrics, including ROUGE-1/2/L\,\cite{lin2004rouge} and BERTScore\,\cite{zhang2019bertscore}, across three evaluation dimensions and multiple domains. We compare these results with the LLM-based method, FineSurE \,\cite{song2024finesure}, an LLM-based evaluation approach implemented using GPT-4o.

Our analysis reveals distinct performance patterns across evaluation dimensions. For faithfulness evaluation, conventional metrics exhibit weaker correlations with human judgments compared to FineSurE across all domains and languages. While completeness and conciseness exhibit stronger correlations than faithfulness, their performance varies depending on the domain.

Nevertheless, conventional metrics generally demonstrate significantly lower agreement with human scores across all dimensions compared to the LLM-based evaluator.


\section{Human Annotation Details}
\label{sec:appendix_annotation}
\paragraph{Qualifications and Compensation}
MTurk annotators are screened through an English proficiency test simulating the fact verification and key-fact alignment tasks. They must also demonstrate a reliable track record at MTurk with a minimum 90\% approval rate and 500 accepted HITs. These crowd-sourced annotators receive compensation exceeding the U.S. minimum wage.

For manual annotation (see Section~\ref{sec:comparsion_uni}), we recruit postgraduate students with proficient English skills (above C2 level) as expert examiners. Specifically, for English-Chinese translation tasks (see Section~\ref{sec:pipeline}) and for any manual annotation involving Chinese, we require the expert examiners to be native Chinese speakers. These experts are compensated at rates exceeding \$30 per hour plus performance-based incentives.

\paragraph{Annotation of Chinese Summaries}
To annotate Chinese summaries, the annotators are provided with summaries translated into English using GPT-4o. These translations undergo our standard process: initial LLM-based translation, validation, and final verification by native Chinese examiners, as described in Section~\ref{sec:input_text_source}. The annotators then work with these verified English translations.

\paragraph{Quality Control}
In addition, since MTurk is a crowd-sourcing platform, it is essential to systematically filter unreliable answers from the annotators. For each summary annotated, we designate about 5-10\% of the annotation unit in a Human Intelligence Task as attention checks, where the correct answers are already known to us. We exclude all responses that do not pass the attention checks. This approach ensures that the annotations collected from MTurk meet the required standards.



\end{document}